\documentclass{article}
\newcommand{\earthsight}{\textsc{EarthSight}}
\usepackage{tikz}
\usepackage{amsmath}
\usepackage{amsfonts}
\usepackage{algorithm}     % Provides the floating "algorithm" environment
\usepackage{algpseudocode} % Provides the pseudocode commands (\State, \For...)
\usepackage{svg}
\usepackage{filecontents}
\usepackage{pgfplots}
\usepackage{booktabs}
\usepackage{multirow}
\usepackage[breaklinks]{hyperref}
\usepackage{tabularx}
\usepackage{float}
\usepackage{makecell}
\usepackage{tikz-cd}
\usepackage{pgfplots}
\usepackage{enumitem}
\usepackage{subfigure}
\usepackage{balance}
\usepackage{microtype}
\usepackage{graphicx}
\usepackage{booktabs} % for professional tables
\usepackage{adjustbox}

\newif\ifshowrevisions
\showrevisionsfalse      % <-- set to \showrevisionsfalse to remove coloring

\ifshowrevisions
  \newcommand{\revised}[1]{\textcolor{blue}{#1}}
\else
  \newcommand{\revised}[1]{#1}
\fi

\usepackage[accepted]{mlsys2025} 

\mlsystitlerunning{\earthsight: A Distributed Framework for Low-Latency Satellite Intelligence}

\usepackage{eso-pic}

\newcommand{\AddBadges}{
  \AddToShipoutPictureFG*{ 
    \put(\LenToUnit{0.65\paperwidth},\LenToUnit{1.23\paperwidth}){ 
      \begin{tabular}{cc} % 'cc' puts them side-by-side
        \includegraphics[width=2.0cm]{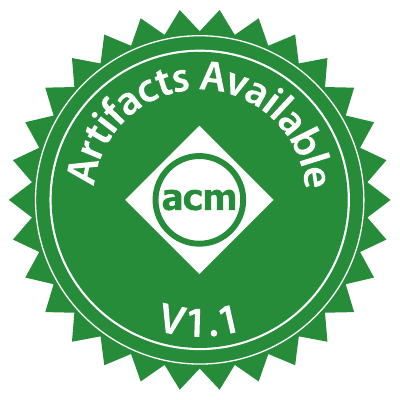} & 
        \hspace{-0.4cm} 
        \includegraphics[width=2.0cm]{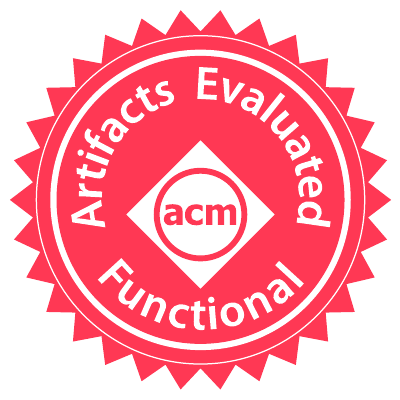}
      \end{tabular}
    }
  }
}

\begin{document}

\AddBadges

\twocolumn[
\mlsystitle{\earthsight: A Distributed Framework for \\ Low-Latency Satellite Intelligence}

% unless you've provided the [accepted] option to the mlsys2025


\mlsyssetsymbol{equal}{*}

\begin{mlsysauthorlist}
\mlsysauthor{Ansel Erol}{GT}
\mlsysauthor{Seungjun Lee}{GT,KAIST}
\mlsysauthor{Divya Mahajan}{GT}
\end{mlsysauthorlist}

\mlsysaffiliation{GT}{Georgia Institute of Technology}
\mlsysaffiliation{KAIST}{KAIST, Daejun, South Korea}

\mlsyscorrespondingauthor{Ansel Erol}{aerol3@gatech.edu}

% % You may provide any keywords that you
% % find helpful for describing your paper; these are used to populate
% % the "keywords" metadata in the PDF but will not be shown in the document
\mlsyskeywords{Computer Vision, Orbital Edge Computing, Satellites, Adaptive Scheduling, MLSys}

\vspace{1ex}

\begin{abstract}
Low-latency delivery of satellite imagery is essential for time-critical applications such as disaster response, intelligence, and infrastructure monitoring. However, traditional pipelines rely on downlinking all captured images before analysis, introducing delays of hours to days due to restricted communication bandwidth. To address these bottlenecks, emerging systems perform onboard machine learning to prioritize which images to transmit. However, these solutions typically treat each satellite as an isolated compute node, limiting scalability and efficiency. Redundant inference across satellites and tasks further strains onboard power and compute costs, constraining mission scope and responsiveness. We present \earthsight, a distributed runtime framework that redefines satellite image intelligence as a \emph{distributed decision problem} between orbit and ground. \earthsight~introduces three core innovations: (1) \emph{multi-task inference} on satellites using shared backbones to amortize computation across multiple vision tasks; (2) a \emph{ground-station query scheduler} that aggregates user requests, predicts priorities, and assigns compute budgets to incoming imagery; and (3) \emph{dynamic filter ordering}, which integrates model selectivity, accuracy, and execution cost to reject low-value images early and conserve resources.
\earthsight~leverages global context from ground stations and resource-aware adaptive decisions in orbit to enable constellations to perform scalable, low-latency image analysis within strict downlink bandwidth and onboard power budgets. Evaluations using a prior established satellite simulator show that \earthsight~reduces average compute time per image by 1.9$\times$ and lowers 90th percentile end-to-end latency from first contact to delivery from 51 to 21 minutes compared to the state-of-the-art baseline.
\end{abstract}

]
% Reduce space above/below figures and captions
\setlength{\textfloatsep}{8pt plus 1.0pt minus 2.0pt}
\setlength{\floatsep}{8pt plus 1.0pt minus 2.0pt}
\setlength{\intextsep}{8pt plus 1.0pt minus 2.0pt}
\setlength{\abovecaptionskip}{4pt plus 1pt minus 1pt}
\setlength{\belowcaptionskip}{0pt plus 1pt minus 1pt}



\printAffiliationsAndNotice{}  % leave blank if no need to mention equal contribution
% \printAffiliationsAndNotice{} % otherwise use the standard text.
% \vspace{-5\baselineskip} 
\enlargethispage{0.5\baselineskip} 

\section{Introduction}

Nanosatellites in Low Earth Orbit (LEO) have transformed our ability to monitor Earth~\cite{mcdowell2020, cbo2023}. 
Commercial constellations such as Planet's Dove and Spire Global's LEMUR now deliver high-resolution, near-daily imagery of the planet’s surface~\cite{dove, lemur}. This data enables a wide array of applications, from environmental monitoring and disaster response to agriculture, urban planning, and intelligence~\cite{thangavel2023, tahir2022, bandarupally2020, nguyen2020, ouchra2022}.
The primary bottleneck in the current Earth observation systems has shifted from data acquisition to analysis covering 200 million km$^2$ per day~\cite{PlanetConstellations}. Thus, to extract timely insights, the community has turned to machine learning~\cite{adagt, barmpoutis2020, tuia2024, wang2025}.
 
However, existing pipelines, wherein all captured imagery is downlinked to ground stations before analysis, introduce substantial delays, often stretching from several hours to multiple days~\cite{serval, l2d2}. Such delays are detrimental for time-sensitive queries: each lost minute can hinder search efforts during natural disasters, obscure evolving damage and environmental conditions, or exacerbate operational uncertainty in conflict zones.
Delays arise from the short, intermittent transmission windows available for downlink, typically 10–15 minutes to offload tens of gigabytes of imagery~\cite{l2d2}.
When satellites cannot assess the analytical value of captured data, images are transmitted in a first-in, first-out manner, potentially causing critical content to wait behind less relevant data.
Planet Labs’ Analytics product shows that while users can run predefined object detection algorithms, the results often arrive days later, making same-day analysis infeasible~\cite{planetanalytics2021}.

\begin{figure}
    \centering
    \includegraphics[width=0.75\linewidth]{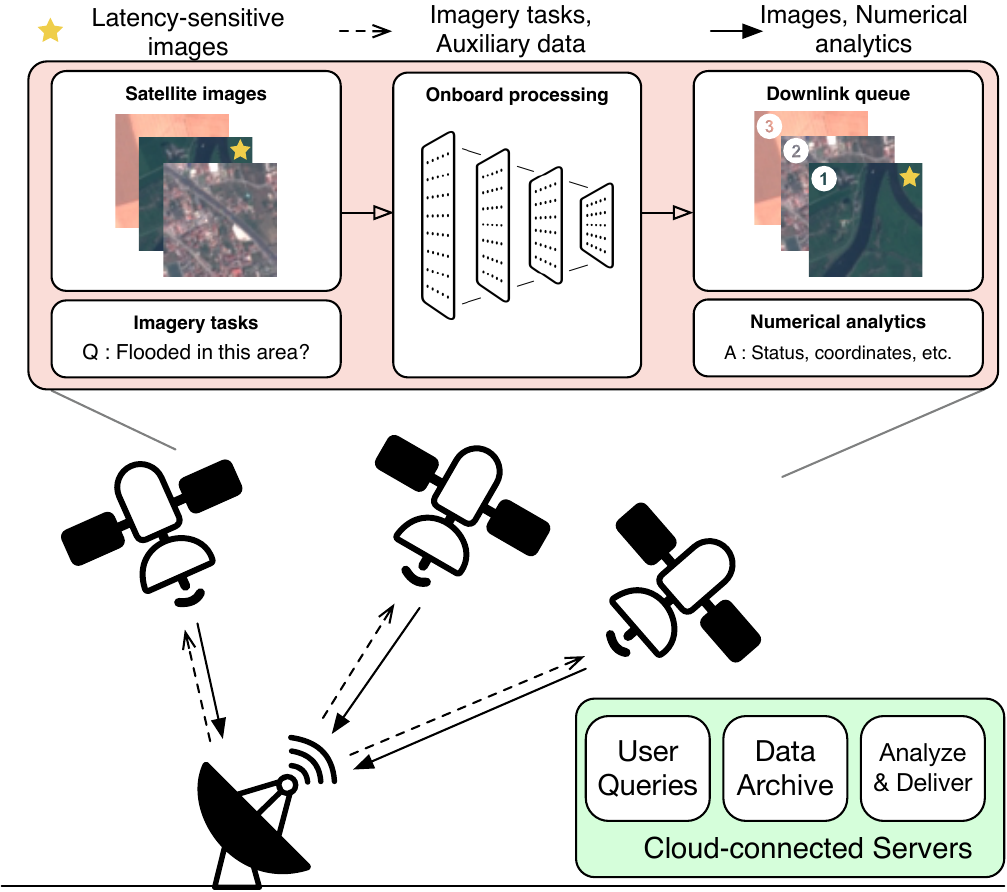}
    \caption{LEO Earth observation systems perform on-board inference for image analysis. Ground stations send tasks generated from user queries to satellites. The satellites prioritize transmission of images identified as latency-sensitive via ML inference.}
    \label{fig:fig1}
\end{figure}

Recent efforts have proposed performing inference on the satellites to reduce latency and conserve bandwidth~\cite{kodan, serval, 10697471, orbitaledge}.
By analyzing images in orbit between capture and downlink, satellites can prioritize, compress, or discard data before transmission (Figure~\ref{fig:fig1}).
However, these approaches treat each satellite as an \emph{isolated node}, ignoring the broader constellation context that governs downlink contention and bandwidth.
Without incorporating global coordination, such designs cannot optimally allocate limited communication opportunities across multiple satellites.

\pagebreak
Moreover, existing systems only support single-task operation. In practice, an image may support multiple concurrent applications, such as ship detection, wake analysis, and illegal fishing identification at sea.
Executing independent inference pipelines for each task is expensive in both computation and energy, leading to redundancy that is unacceptable in resource-constrained environments.
The multi-task nature of satellite workloads and single-purpose isolated design of prior systems limits their scalability and responsiveness.

\emph{\earthsight~redefines image analysis as a distributed decision problem spanning both ground and orbit.
Rather than viewing satellites as passive data forwarders or independent inference engines, we offer a joint scheduling framework that leverages global context from the ground station with local resource-aware adaption in-orbit.}

Enabling \earthsight~requires overcoming a variety of challenges. Nano-satellites operate under tight power budgets, harvested via small solar panels and used for imaging, communication, and control systems. As a result, only a fraction of power is available for compute, forcing the use of low-power, radiation-hardened processors that lag far behind terrestrial accelerators in capability.
Consequently, onboard image processing often cannot complete before the next downlink window, limiting responsiveness for time-sensitive missions.
To address these challenges, we employ an integrated approach that combines multi-task learning with globally informed pre-processing at the ground station.
This allows the satellite constellation to collaboratively manage diverse, time-critical workloads within limited compute and bandwidth envelopes.
To facilitate the distributed decision, \earthsight~proposes three coordinated components: (1) multi-task onboard inference for priority annotation, (2) query schedule construction at ground stations, and (3) adaptive filter ordering on satellites.
\vspace{1 em}
\earthsight~employs a multi-task model consisting of shared feature backbones and task-specific heads, amortizing inference costs across multiple filters without compromising on model size or capacity.
While this design improves efficiency, it introduces a dependency: task heads can only execute after the backbone’s latent features are computed.
\earthsight’s scheduler resolves these dependencies by adjusting filter orderings at runtime, jointly optimizing for precision, latency, and available power.

At the ground station, \earthsight~analyzes historical image distributions and mission policies to synthesize a preliminary schedule. For each expected capture, it aggregates filters from all active user queries, e.g., cloud-free and has fire or flooding near a populated area, into compact logical formulas. It then simulates future transmission windows to forecast two key parameters: the minimum image priority likely to be transmitted and the expected compute budget per image. These parameters are integrated into the computation schedule to supply the global context that guides approximate, truncated inference onboard the satellite.

At runtime, satellites execute ML filters to prioritize images. To manage power constraints, they employ a model-ordering strategy that stops as soon as an adaptively-set confidence threshold is met. The sequence of filters is critical; if two ML-based filters have comparable runtimes, the one more likely to influence the prioritization outcome should execute first. 
Finding an optimal sequence of filter operations for each image is crucial for efficiency, however is \textit{NP-}hard and thus computationally intractable.
\earthsight~addresses this with a heuristic approach that defines each filter’s utility as a function of four factors: (i) the filter's accuracy (ii) the filter's execution time (cost), (iii) the probability of a negative result, based on prior information from the ground station, and (iv) the influence of the filter's outcome on the overall query formula.
This cost–benefit formulation enables the scheduler to repeatedly select the next filter most likely to yield a rapid decision on the image.

\noindent In summary, \earthsight~contributes:
\begin{enumerate}[leftmargin=*, itemsep=1pt, parsep=0pt, topsep=1pt, partopsep=1pt]
\item A \textbf{distributed decision framework} that redefines onboard image analysis as a joint process between ground and orbit, merging global context from ground stations with local, resource-aware adaptation in satellites.
\item \textbf{Multi-task onboard inference} that amortizes computation across multiple application-specific tasks through a shared backbone, enabling efficient, scalable analysis under tight power budgets.
\item \textbf{Adaptive filter scheduling} based on a utility-driven ordering that integrates selectivity, execution time, accuracy, and logical impact to maximize early rejection of low-priority images.
\end{enumerate}

\noindent Through hardware-enhanced simulation studies on three multi-purpose scenarios, we demonstrate that \earthsight~ alleviates the computational bottleneck, resulting in a decrease in the 90th percentile tail latency from first contact to image delivery from 51 to 21 minutes relative to the state of the art baseline, \textsc{Serval}~\cite{serval}.

% \begin{flushleft}
% Our code is open-source and available at 
% {\footnotesize{\href{https://github.com/earthsight-contributors/earthsight}{\texttt{https://github.com/earthsight-contributors/earthsight}}}}.
% \end{flushleft}

%---------------------------------------------------------------------
\section{Background}
%---------------------------------------------------------------------

Satellites in Low Earth Orbit (LEO) offer frequent, high-resolution, multi-spectral imaging of Earth’s surface, typically completing a full orbit every 90 minutes at a few hundred kilometers altitude.
While early Earth observation (EO) relied on large, monolithic platforms \cite{sentinel_data, aqua_cer_ssf}, the past decade has seen a shift towards deploying nanosatellites, compact spacecraft ranging from 1000 to 8000 cubic centimeters \cite{SpireConstellation, PlanetConstellations}. These  satellites, equipped with commercial off-the-shelf (COTS) components such as the NVIDIA Jetson series, form a globally distributed computing platform capable of onboard inference.

\paragraph{Orbital Edge Computing.} The concept of processing data directly on satellite, i.e., Orbital Edge Computing (OEC), conserves downlink bandwidth by filtering out unusable data, such as cloudy images, prior to transmission~\cite{orbitaledge}. The idea now supports commercial approaches, like those pursued by Spire Global under contracts with NASA and the Canadian Space Agency, to develop onboard wildfire detection using Convolutional Neural Networks \cite{spire2025wildfiresat}.  

\revised{Recent initiatives have explored more powerful compute substrates for space-based inference: Google's Project Suncatcher investigates deploying TPUs in orbit to enable large-scale ML workloads~\cite{suncatcher2025}, while platforms like Pelican have demonstrated Jetson-based deployments for real-time onboard processing~\cite{pelican}. These developments underscore the growing feasibility of sophisticated edge inference in LEO environments.}

\paragraph{Leveraging Onboard Processing for Image Prioritization.}
The push toward onboard processing is driven in part by the constraints of satellite systems compared to ground-based infrastructure drawing on data centers and cloud resources.
In orbit, compute capability is fixed at launch, power-limited, and must support all imaging, communication, and control operations within a 1 to 5 watts per unit budget~\cite{powerbudgets}. These constraints prevent continuous, high-throughput processing and limit how many models can run per image.
To mitigate this, recent hybrid systems~\cite{serval, kodan, 10697471} split computation between satellites and ground stations, estimating image utility and dynamically compressing and reordering transmissions based value or urgency.

At the same time, satellites face brief downlink windows and constrained communication bandwidth \cite{so2022fedspace}. 
Thus, these aforementioned onboard ML-based pipelines prioritize high-utility images, those showing disasters, flooding, or anomalous land use—while deprioritizing or discarding less informative data. Each image is typically processed by multiple models targeting distinct attributes to build a comprehensive understanding of the scene.
Because satellite images are large and high-resolution, they are tiled before inference to avoid downsampling artifacts~\cite{orbitaledge}. Modern systems further leverage contextual cues such as power availability, cloud cover, and location metadata to decide when to run full inference, skip processing, or use lightweight approximations (e.g., downsampled inputs)~\cite{kodan, serval}.

%\paragraph{Multi-Task Models for Onboard Inference.}
%
% Most existing OEC systems perform single-task inference per image~\cite{orbitaledge, kodan, serval} which simplifies deployment but underutilizes the rich information in satellite imagery. In reality, each image can serve multiple downstream tasks, and joint processing can improve both efficiency and utility. Multi-task learning addresses this by sharing feature extraction across tasks, reducing memory overhead, and supporting multiple mission goals without deploying multiple large models~\cite{caruana1997}. These models use a shared backbone to extract latent features, followed by lightweight task-specific heads that leverage general-purpose representations~\cite{yu2024, daroya2024}. 
\paragraph{Multi-Task Models for Onboard Inference.}
Most existing OEC systems perform single-task inference per image, running isolated pipelines to identify a specific feature such as cloud cover or vessels~\cite{orbitaledge, serval, kodan}. While this simplifies deployment, it underutilizes the rich information in satellite imagery. In reality, each image can serve multiple downstream tasks, and joint processing can improve both efficiency and utility. Multi-task learning addresses this by sharing feature extraction across tasks, reducing memory overhead, and supporting multiple mission goals without deploying multiple large models~\cite{caruana1997}. These models use a shared backbone to extract latent features, followed by lightweight task-specific heads that leverage general-purpose representations~\cite{yu2024}. For example, \citet{daroya2024} successfully demonstrated this in remote sensing by replacing five separate single-task models with a unified multi-task network to simultaneously generate masks for water, clouds, and shadows, significantly reducing computation time.

Multi-task neural nets typically adopt either \emph{hard} parameter sharing, with a single backbone and multiple heads, or \emph{soft} parameter sharing, where separate models are trained with coupling constraints ~\cite{ruder2017}. As the former offers greater computational efficiency~\cite{cross}, \earthsight~employs hard parameter sharing to minimize inference costs for diverse applications.

Building on these foundations, \earthsight~defines onboard image analysis as a \textit{distributed decision framework that unifies global scheduling at the ground station with local, power-aware adaptation in orbit}, enabling scalable, multi-task image analysis under tight compute, bandwidth, and latency constraints.
\section{\earthsight~System}

\begin{figure}[t]
    \centering
    \includegraphics[width=1\linewidth]{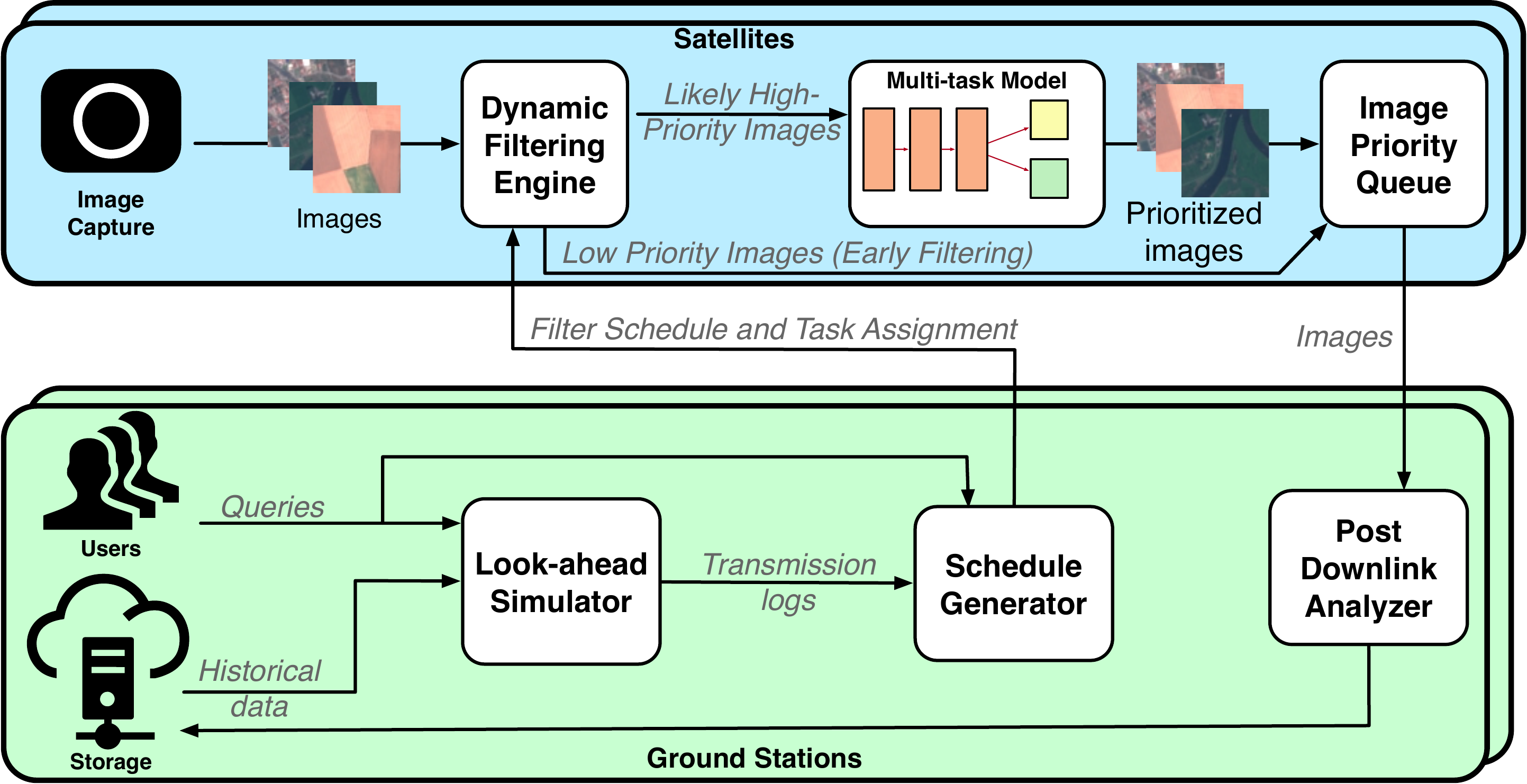}
    \caption{Overview of the \earthsight~system. \earthsight~integrates onboard multi-task models with predictive scheduling to enable query-driven, low-latency analysis of satellite images.}
    \label{fig:earthsight}
\end{figure}

\earthsight~frames onboard image analysis as a \emph{distributed decision problem} spanning ground and orbit, coordinating \textit{global context} from ground stations with local, \textit{resource-aware adaptation} on satellites. Unlike prior systems that treat satellites as isolated inference nodes, \earthsight~dynamically selects and schedules tasks based on available power, predicted image utility, and mission context.  
The pipeline, shown in Figure~\ref{fig:earthsight}, integrates three key components:  
(1) \textbf{Multi-task models} that perform multiple vision tasks per image to reduce redundant computation;  
(2) \textbf{Ground-station scheduling} that uses auxiliary metadata, such as location, orbit timing, and power forecasts, to generate predictive computation plans for satellites; and  
(3) An \textbf{in-orbit runtime} that dynamically orders and executes filters to minimize expected compute time while prioritizing images for downlink.  
\revised{Like prior work~\cite{serval}, \earthsight~does not \textit{discard or drop any images} and only optimizes the transmission order to maximize responsiveness.} Together, these components form a flexible, resource-aware system that adapts to changing mission conditions while maximizing the value of each downlinked image.

\subsection{Multi-Task Neural Architecture}
\label{system:mtl}

\begin{figure}[t]
    \centering
    \includegraphics[width=0.95\linewidth]{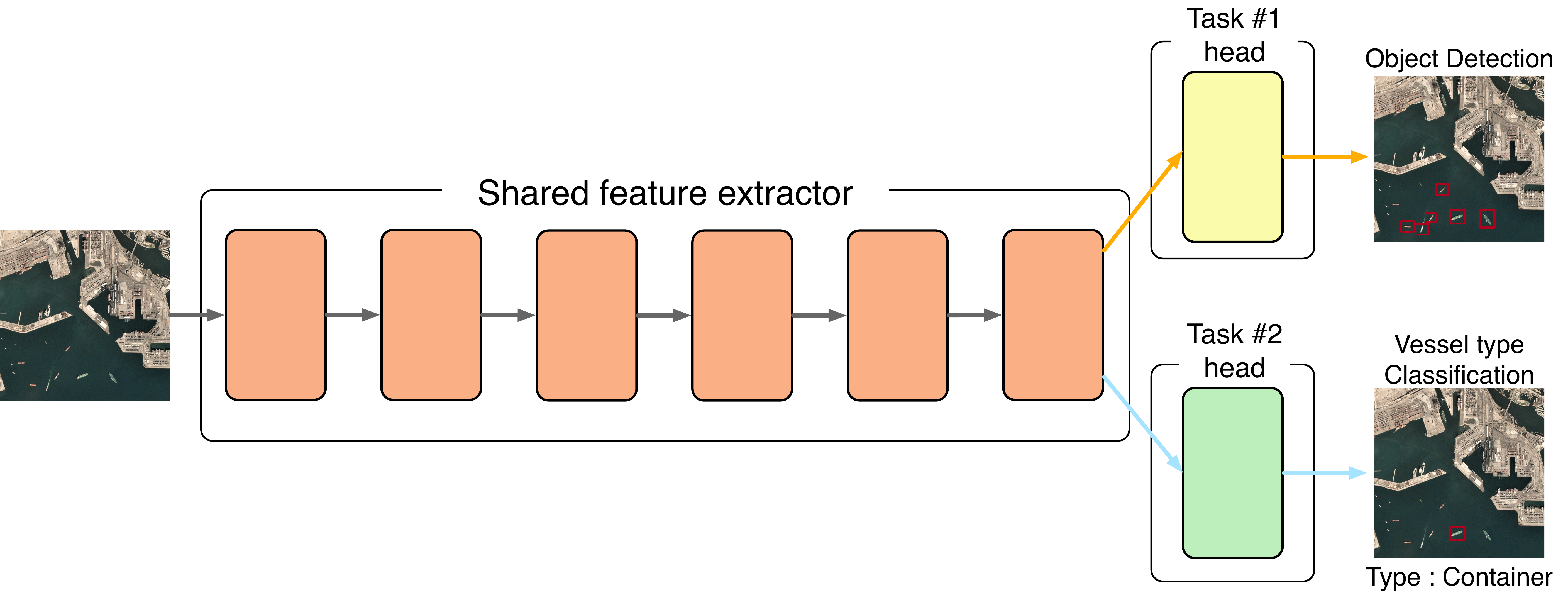}
    \caption{\earthsight’s multi-task model architecture. A shared backbone processes the input image to produce a latent representation, which is used by lightweight, task-specific heads to perform heterogeneous tasks such as classification.}
    \label{fig:mtl}
\end{figure}

\revised{\earthsight~employs multi-task learning to balance two fundamental and conflicting requirements: the need for depth to accurately detect anomalies and events, and the need for efficiency on power-limited edge hardware. By sharing computation through common backbones and attaching task-specific heads, \earthsight~minimizes total evaluation cost without sacrificing representational capacity. This approach amortizes backbone costs while maintaining features present only in deeper layers, yielding deterministic latency and a small footprint, key properties for Orbital Edge Computing tasks.}

\earthsight~supports multi-task models that follow the standard paradigm of a shared backbone and task-specific heads (Figure~\ref{fig:mtl}). A model maps input image $x$ to multiple outputs $\{y_{t}|t\in T\}$, where $T$ denotes the tasks that the model supports. However, \earthsight~introduces two distinct architectural modifications compared to conventional Earth observation multi-task networks (e.g., \citet{daroya2024}). 

First, while standard multi-task models execute a single monolithic forward pass to evaluate all heads simultaneously, \earthsight~architecturally decouples head execution. The computational graph is structured so the runtime can extract the latent representation $z$ via $f_{\text{backbone}}(x) = z$, and then conditionally and sparsely evaluate only the required task heads $f_{\text{head}}^{t}(z) = y_t$ based on the dynamic schedule. This lightweight design allows multiple tasks to share core feature computations while preserving task-level specialization, striking a balance between memory efficiency, task-specific performance, and limited satellite resources.

Second, rather than forcing all tasks onto a single backbone, \earthsight~partitions tasks into domain-specific clusters, assigning a dedicated EfficientNet backbone to each group, coupled with lightweight multi-layer perceptrons as prediction heads. The architecture reflects real-world satellite needs as the relevant features to analyze at maritime versus in desert terrain exhibit strong intra-group similarity but have little interaction between groups. This clustering allows the backbone to capture generalizable features within a domain, while keeping model sizes small enough to fit edge AI accelerators. Attempting to merge these heterogeneous domains into a single monolithic model would either underutilize shared representations or exceed satellite memory limits~\cite{zhang2018}.

\revised{In designing this architecture, we also considered several alternatives but found them less suitable for Earth-observation workloads. Early-exit mechanisms~\cite{earlyexit} incur the memory overhead of loading the full backbone, while at the same time miss deeper, fine-grained features and exhibiting highly variable latency. Similarly, Mixture-of-Depths models~\cite{raposo2024mixtureofdepthsdynamicallyallocatingcompute} introduce storage and runtime decision overheads without overlapping compute. Among the available model frameworks, we found multi-task learning to be the most effective at achieving model depth in an efficient way. Section \ref{sec:related-work} discusses further alternatives for model-level optimization.}

\earthsight~is designed for scenarios in which multiple tasks are performed on each input image. By overlapping backbone computations across tasks, the multi-task architecture maximizes inference throughput relative to single-task deployments. Single-task models are naturally supported as a special case of the multi-task framework, showcasing the flexibility and scalability of \earthsight. \revised{Furthermore, the modular design simplifies how \earthsight~handles new tasks, presenting operators with four options: (1) reusing an existing, compatible backbone, (2) adding the task stand-alone, (3) fine-tuning backbones and heads offline without re-clustering, or (4) re-clustering and training new backbones. Options 1 and 2 allow for immediate deployment, while 3 and 4 can be prepared in the background for optimized performance. \earthsight's modularity enables lightweight uplinks of new task heads without transmitting full backbone parameters, addressing bandwidth limitations.}

\subsection{Scheduler at the Ground Station}
\label{system:ground_scheduler}
The \earthsight~scheduler, resident on the ground station, orchestrates all information required for the satellite runtime to dynamically route inference tasks and prioritize imagery. This includes auxiliary metadata such as updated model weights, task-specific success probabilities, and the precomputed task schedule. Offloading this decision logic on the ground allows \earthsight~to leverage abundant compute and memory resources, which are infeasible on power- and compute-constrained nanosatellite.

\paragraph{Query-driven framework.}
\earthsight~adopts a query-centric design where downstream applications define task requirements in terms of latency sensitivity and image selection criteria. Some high-priority queries (e.g., Planet’s Tasking Product~\cite{planet_tasking}) demand immediate downlink of specific regions, bypassing onboard inference. Others, such as segmentation or classification tasks, require onboard processing to reduce bandwidth usage while still enabling timely decision-making.
For latency-sensitive inference tasks, the scheduler is designed to minimize total computation while prioritizing correctness: it is preferable to occasionally transmit unimportant images (false positives) than to delay or omit critical data (false negatives). 

To formalize this, \earthsight~adapts the boolean query syntax introduced by \citet{serval}. While Serval formulates user queries as logical expressions over visual attributes (e.g., ``cloud-free AND ship present AND ship is military'') within an Area of Interest (AOI) primarily to bifurcate static and dynamic filtering tasks, \earthsight~retains this foundational logical structure but significantly extends the interface to support distributed, multi-task orchestration. Specifically, we augment each query with a priority level $p \in \{1,2,3,4,5\}$. This adaptation allows the ground station to aggregate overlapping logical conditions from multiple downstream applications and compile them into a unified execution schedule that encodes task selection, dynamic priority thresholds, model updates, and expected filter odds.

\paragraph{Look-ahead simulation.}
Given the temporal variability of downlink windows and bandwidth distribution across satellites, \earthsight~integrates a look-ahead simulator at the ground station. This simulator predicts satellite behavior, anticipated data volume, and the prioritization of bytes transmitted per downlink window. Priority-weighted byte counts account for the probabilistic nature of image importance; for instance, a 50KB image with a 10\% chance of being each priority contributes 10KB to each priority bin.
To address potential misclassifications by onboard filters, an intermediate priority level $p_{compute}$ is introduced between priorities $p = 1$ and $p = 2$. Images processed onboard but rejected are transmitted within this level, ensuring they are delivered before guaranteed low-priority content. 
% This mechanism balances the need for bandwidth efficiency with robust coverage of potentially critical imagery.

The simulator also computes two key quantities for each satellite: the lowest priority threshold $p^*$ for which all images above it can be transmitted in the next downlink window, and the rejection rate $r_{\text{reject}} = \frac{|\text{inference}| - |\text{downlink}|}{|\text{inference}|}$, which quantifies the proportion of computed images deprioritized to guarantee timely delivery of higher-priority content. Because satellites may have hours between contacts, this look-ahead analysis is crucial to guarantee responsiveness for time-sensitive tasks. When ground-truth information becomes available, the simulation is adjusted accordingly, and the resulting changes are propagated through the system. Although this update process is computationally expensive, our Python implementation is optimized to perform simulations spanning several hours in $\mathcal{O}(\text{minutes})$.

\paragraph{Schedule generation.}
Upon contact, satellites request an updated task schedule. \revised{The ground station maps the satellite’s planned capture locations to the set of queries whose AOIs intersect each image using an off-the-shelf R-tree index, which bounds AOIs in rectangular extents for efficient spatial search. For each image, it joins the filtering criteria of retrieved queries to constructs a Disjunctive Normal Form (DNF) formula} encoding the conditions under which the image exceeds the computed threshold $p^*$.

In addition, the ground station maintains up-to-date filter success probabilities and lightweight model weight updates, derived from historical datasets (e.g., weather, incident reports) or prior inference results. This provides partial, probabilistic hints about the contents of each image to the satellite runtime, in the form of task schedules, filter success probabilities, and lightweight model updates. This enables the onboard runtime to make dynamic, context-aware decisions, performing inference only when necessary to prioritize images efficiently within the satellite’s tight compute and power constraints.
This coordination between ground and orbit showcases \earthsight’s distributed decision framework, allowing near-optimal image prioritization under tight compute, power, and bandwidth constraints.

\subsection{In-Orbit Satellite Runtime}
\label{system:sat_runtime}

The satellite runtime is responsible for executing prioritization tasks efficiently while maintaining high decision quality. Once an image is captured that requires inference according to the ground-generated inference schedule (Section~\ref{system:ground_scheduler}), the in-orbit runtime iteratively evaluates filters until the image is prioritized. 
Within \earthsight's distributed framework, this runtime forms the \emph{onboard inference layer}, complementing \emph{ground-side scheduling}. The ground scheduler provides a predictive execution plan that balances utility and resource constraints across satellites, while the onboard runtime refines these decisions dynamically using real-time power and inference outcomes. 
This co-design ensures that satellite-side decisions have hints from ground stations thus can remain locally optimal yet globally consistent, connecting onboard adaptivity to the distributed scheduling policy.

\paragraph{Adaptive Prioritization Policy.}
To manage limited onboard resources, the runtime employs adaptive upper ($\alpha$) and fixed lower ($\beta$) confidence thresholds that guide priority decisions based on current power and target rejection rates. Filter execution order is dynamic based on a utility function (Equation~\ref{eq:utility}), balancing speed, informativeness, and correctness. Progress toward a prioritization decision is tracked using a confidence score (Equation~\ref{eq:confidence}), estimating the probability that the image satisfies the prioritization criteria given the filters executed so far. 

Images are assigned priority $p_{compute}$ if the likelihood of satisfying their formula is below \(\beta\), and the maximum priority of a remaining term if that likelihood exceeds \(\alpha\).
Because the cost of incorrectly rejecting a valuable image is high, \(\beta\) is fixed to a small value, while \(\alpha\) is dynamically adjusted at runtime. Specifically, \(\alpha\) evolves according to:
\begin{equation}
\footnotesize
\alpha_t = \min(1, \alpha_{t-1} + \lambda_1(r_{\text{power}, t} - 1) + \lambda_2(r_{\text{dep}, t} - r_{\text{reject}})) 
\end{equation}
\normalsize
where \(\alpha_t\) is the threshold at time \(t\); \(\lambda_1, \lambda_2\) are tuning parameters; \(r_{\text{power}, t}\) is the ratio of current to target power level (set to 70\% of maximum charge); \(r_{\text{dep}, t}\) is the fraction of computed images identified as low-priority); \(r_{\text{reject}}\) is the target rejection rate from Section~\ref{system:ground_scheduler}, and  $\min(1,\cdot)$ caps $\alpha$ at 1.
This adaptive threshold allows the satellite to relax its selectivity when power is scarce or bandwidth is limited, ensuring robust operation across varying conditions.

\begin{figure*}[h]
\centering
\makebox[0.85\textwidth][c]{%
\includegraphics[width=0.9\textwidth]{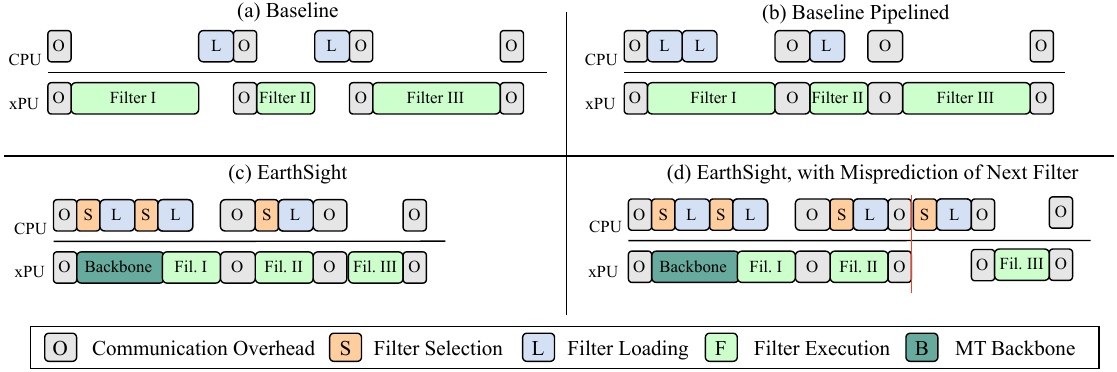}%
}

\caption{Timing diagrams illustrating filter execution strategies on a CPU-xPU system: (a) Serval's (baseline) sequential execution~\cite{serval}, (b) Pipelined baseline with overlapped loading, (c) \earthsight~with predictive filter selection and loading, and (d) \earthsight~with a filter misprediction. The phases represent Filter Selection (runtime scheduling), Filter Loading (CPU model preparation), Filter Execution (xPU inference), and Communication Overhead (CPU-xPU coordination).}

\label{fig:running_seq}
\end{figure*}

\paragraph{Utility-Driven Filter Ordering for Early Rejection.}
To reduce onboard computation, the runtime aims to reject unpromising images early by dynamically ordering filters according to their expected utility. Each filter \(f_i\) has an associated time cost $t(f_i)$ and likelihood of evaluating true $p_i$. To incorporate multi-task models, we define $b_i$ as the shared computational backbone (e.g., a neural network feature extractor) required by \(f_i\). Let $t(b_i)$ be the cost of executing this backbone. The effective execution time is:
\begin{equation}
\label{eq:t_eff}
\footnotesize
t_{\mathrm{eff}}(f_i, \mathcal{E}) =
\begin{cases}
t(f_i), & \text{if } b_i \in \mathcal{E} \text{ or } b_i = \emptyset \\
t(f_i) + t(B(f_i)), & \text{otherwise}
\end{cases}
\normalsize
\end{equation}
where $\mathcal{E}$ denotes the current execution state.  

The overall goal is to find an execution policy \(\pi^*\) that determines the sequence of filters to evaluate for each image to minimize the expected total execution cost required to reach a decision boundary. The objective is to find the policy \(\pi^*\) that minimizes:
\vskip -3 mm
\begin{equation}
\label{eq:objective}
\footnotesize
\mathbb{E}[C(\pi)] = \mathbb{E}\!\left[\sum_{j=1}^{K} t_{\mathrm{eff}}(f_{(j)}, \mathcal{E}_{j-1})\right]
\normalsize
\end{equation}

where \(K\) is the number of filters evaluated before the confidence exits $[\beta,\alpha]$.  

Finding the optimal policy \(\pi^*\) is an instance of the Stochastic Boolean Function Evaluation (SBFE) problem, which is \textit{NP}-hard (see Section \ref{sec:related-work}). \revised{Exact methods would take years per image and are thus computationally intractable for real-time operations. We overcome this via a greedy strategy that, at each step, selects the filter maximizing the immediate information gain per unit of execution time to approximate Equation \ref{eq:objective}, as is typical for submodular optimization problems such as SFBE~\cite{golovin2011adaptive}.}

\paragraph{Utility Function.}
The utility of a filter \(f_i\) given a partial execution state \(\mathcal{E}\) is:
\begin{equation}
\label{eq:utility}
\footnotesize
U_\phi(f_i, \mathcal{E}) = \frac{(1 - p_i) \cdot \text{tpr}_i \cdot n_i}{t_{\mathrm{eff}}(f_i, \mathcal{E})}
\normalsize
\end{equation}
where \(p_i\) is the pass probability, \(\text{tpr}_i\) is the true positive rate, and \(n_i\) is the number of DNF terms containing \(f_i\). 
Filters with higher \(U_\phi\) are executed earlier, improving early rejection efficiency while maintaining correctness. \revised{Our utility function offers a middle ground between static or naive ordering strategies, which miss opportunities for improving efficiency, and more complex methods like meta-heuristic search that introduce substantial computational overhead.}

\paragraph{Confidence and DNF Completion.}
Given execution state \(\mathcal{E}\), the probability that an image satisfies its prioritization formula is:
\begin{equation}
\footnotesize
C_\phi(\mathcal{E}) = 1 - \prod_{T \in \phi} \left(1 - P(T, \mathcal{E})\right),
\label{eq:confidence}
\normalsize
\end{equation}
where each term \(T\) contributes
\begin{equation}
\footnotesize
P(T, \mathcal{E}) =
\begin{cases}
0, & \exists f \in T \text{ s.t. } \mathcal{E}[f] = \texttt{False}, \\
\prod_{f \in T_u} p_f, & \text{otherwise}
\end{cases}
\normalsize
\end{equation}
and \(T_u = T \setminus \mathcal{E}\) is the set of unevaluated filters.  
The confidence accounts for both known failures and statistical predictions about unevaluated filters. If any term is fully satisfied, the image is accepted early. Otherwise, processing continues until the image is prioritized.

\paragraph{Execution Strategy.}
At each iteration, the filter with the greatest utility $U_\phi(f_i, \mathcal{E})$ is executed, and the execution state $\mathcal{E}$ is updated. If for each term, at least one filter returns \texttt{False}, the image is immediately marked with priority $p_{compute}$ (see Section \ref{system:ground_scheduler}), and no further filters are processed. If all filters in a term of $\phi$ return \texttt{True}, the image is annotated with the highest priority of a satisfied term.

\earthsight~\textit{decouples correctness from efficiency}. While the utility heuristic relies on probabilistic priors ($p_i$), \earthsight~is structurally insulated from estimation errors. Because the lower confidence threshold ($\beta$) is fixed, the runtime never rejects an image probabilistically; demotion to the $p_{compute}$ tier strictly requires the DNF formula to evaluate to \texttt{False}. Conversely, inaccurate priors can only inflate confidence past the dynamic upper threshold ($\alpha$), causing a false positive. Thus, probability drift gracefully degrades execution latency and bandwidth efficiency rather than causing critical data loss, preserving strict logical correctness for negative evaluations.
% \begin{figure}
% \centering
% \includegraphics[width=\linewidth]{./figures/simple_timing_diagram.pdf}%
% \caption{Timing diagrams shows pipelined filter execution on a CPU-xPU system for \earthsight~without mispredictions.}
% \label{fig:running_seq_simple}
% \end{figure}

Finally, the onboard runtime periodically reports telemetry statistics, such as effective rejection ratios, average $\alpha_t$ adjustments, and compute utilization, back to the ground scheduler at downlink. These aggregated metrics allow the ground station to recalibrate model success probabilities and update the look-ahead simulator, thereby refining the next round of scheduling decisions. \emph{This feedback loop completes \earthsight’s distributed decision framework: the ground optimizes long-horizon coarse policy, while offering hints to the satellites to enact and locally adapt those policies in real time in a fine grained manner per satellite, jointly achieving efficient, responsive, and globally consistent image prioritization across the constellation.}

\subsection{Mitigating the Overhead of \earthsight}
\label{system:mitigation}

While \earthsight’s runtime introduces adaptability and responsiveness, it also incurs computational and communication overhead relative to prior proposed static pipelines~\cite{serval, orbitaledge}. To maintain efficiency, we employ two optimizations: compact schedule representation and pipelined CPU–xPU execution.

\paragraph{Schedule Compression.}
Each image corresponds to a DNF formula $\phi$, but storing all formulas individually is impractical, e.g., 45 images per minute for 6 hours yields over 16,000 formulas. In practice, the number of \textit{unique} formulas for one satellite over six hours is small (typically $<256$).  
Leveraging this insight, we compress the schedule by first identifying the unique set of filters, $\Phi_{unique}$. We then construct a lookup table that indexes these formulas from 1 to $|\Phi_{unique}|$, using a single byte per entry. The complete schedule $S = \phi_1, \phi_2, \ldots$ is thus encoded as a lookup table followed by the sequence of 1-byte indices, yielding an average compression of $\sim25\times$. \revised{As a result of these optimizations, the network overhead for distributed coordination is less than 0.1\%, when compared with the $>200$GB of daily image downlink volume.}

\paragraph{Pipelined Runtime and Filter Execution.}

\earthsight~runs models on a satellite xPU (e.g., GPU or TPU) while coordinating decision logic on the CPU. To minimize latency, we pipeline CPU-side preparation with xPU-side execution: while the accelerator evaluates the current filter $F_k$, the CPU prepares the next likely filter $F_{k+1}$. When $F_k$ is long-running, the CPU speculatively prefetches alternate filters ($F_{k+1}^{\text{prob}}$, $F_{k+1}^{\text{alt}}$), hiding decision latency even under mispredictions (Figure~\ref{fig:running_seq}).  
This overlap ensures continuous hardware utilization and reduces idle time.

\emph{Collectively, schedule compression and pipelined execution enable the \earthsight~to meet throughput and latency goals at scale, allowing each satellite to adaptively prioritize images in concert with the global inference schedule.}

\section{Evaluation and Results}
%-------------------------------------------------------------------------------

\earthsight’s performance as a distributed decision framework for scalable, low-latency satellite intelligence is evaluated using \textit{real hardware measurements and trace-driven constellation-scale simulations}, consistent with all related prior work~\cite{serval, kodan, l2d2, 10697471}. Specifically, we deploy the in-orbit runtime on off-the-shelf accelerators, Coral Edge TPU~\cite{coral} and Jetson AGX Orin GPU~\cite{jetsonorin}, to obtain empirical inference times and power profiles. These metrics are then integrated into our simulation framework to model end-to-end latency across an entire satellite constellation.

\subsection{Real-world Scenarios and Multi-task Datasets}

\paragraph{Scenarios Evaluated.} To analyze the performance of our system, we use three real-world application scenarios as benchmarks. See Appendix \ref{app:scenarios} for detailed descriptions, prioritization, and image samples. \textbf{Natural Disaster Monitoring} concerns early warning and impact assessments for floods, wildfires, and earthquakes. The \textbf{Intelligence} scenario constitutes strategic awareness of maritime and aerial activities in geopolitically significant regions, including vessel classification and disambiguating military and commercial traffic. The \textbf{Urban Earth Observation} scenario is a comprehensive monitoring framework for major global cities, integrating disaster (e.g., urban flooding, building integrity, environmental change) and security elements tailored to city-specific risks.

\paragraph{Multi-task Models.}
As part of these scenario evaluations, we assess \earthsight’s ability to support multi-task inference, analyzing trade-offs between memory footprint and prediction accuracy across configurations. The evaluated models are trained on representative datasets spanning our three application scenarios using three classification and three segmentation datasets, including EuroSAT, ADVANCE, and PatternNet~\cite{boguszewski2022, helber2019, zhou2018}. Dataset details are summarized in Table~\ref{tab:dataset_summary_grouped_single}.

\begin{table}[t]
\centering
\caption{Datasets Used for Model Evaluation}
\label{tab:dataset_summary_grouped_single}
\renewcommand{\arraystretch}{1.2}
\resizebox{1\linewidth}{!}{
\begin{tabular}{@{} l l l c p{4.2cm} @{}}
\toprule
\textbf{Task} & \textbf{Dataset} & \textbf{Resolution} & \textbf{\# Classes} & {\textbf{Collection Source}} \\
\midrule
\multirow{3}{*}{\shortstack{Image\\Classification}}
& EuroSAT                 & $\sim$10m       & 10 & Sentinel-2 satellite \\
& ADVANCE                 & 0.5m            & 13 & Google Earth Engine \\
& PatternNet              & Varies          & 38 & {Google Earth Engine} \\
\midrule
\multirow{3}{*}{\shortstack{Semantic\\Segmentation}}
& LandCover.ai            & 25-50cm        & 3  & Aerial Photography (Poland) \\
& LoveDA                  & 0.3m            & 7  & {Google Earth Engine} \\
& DeepGlobe LC    & 50cm            & 7  & DigitalGlobe Nanosatellites \\
\bottomrule
\end{tabular}
}
\end{table}

\subsection{Evaluation Setup}
\paragraph{Hardware Platforms.}
We evaluate our system using two distinct edge AI coprocessors: the Google Coral Edge TPU and the NVIDIA Jetson AGX Orin Nano GPU (Table~\ref{tab:coprocessorspecs}). The Coral delivers a consistent 4~TOPS while consuming only $\sim$2W~\cite{coral}. It functions as a simple, near-instant-on peripheral, making it ideal for low-power, intermittent tasks. The Jetson AGX Orin Nano is a full-featured System-on-Module with Ampere architecture, offering up to 40~TOPS with a configurable power budget starting at 7W \cite{jetsonorin}. \revised{These coprocessors are widely used OEC accelerators across industry, government, and academia, including NASA, Spire, Planet, Lockheed Martin, Sidus Space, and others~\cite{goodwill2021nasa, lockheed2020lajument, pelican, suncatcher2025, hardware2024review}.}
%Its key drawback as a coprocessor is the significant idle power consumption and long start times before the first inference. We optimize our models for the Edge TPU and include Jetson for comparison.

\begin{table}[t]
    \centering
    \caption{Edge AI Coprocessor Specifications}
    \label{tab:coprocessorspecs}

    \small
    \resizebox{0.95\linewidth}{!}{%
    \begin{tabular}{lll}
        \hline
        \rule{0pt}{0.8em}
        \textbf{Specification} & \textbf{Google TPU} & \textbf{NVIDIA GPU}\\
        \hline
        FLOPs & 4 TOPS (INT8) & 40 TOPS (INT8) \\
        Processor & Coral Edge TPU & Jetson AGX Orin Nano\\
        CPU & Quad-core Arm A53 & 6-core Arm A78AE \\
        Memory & 1 GB LPDDR4 & 8 GB LPDDR5 \\
        Power & 2W (peak) & 7W -- 15W\\
        \hline
    \end{tabular}%
    }
\end{table}

\paragraph{Simulating the Satellite Constellation.}
We verify our results at scale using a Python-based simulator that extends the validated open-source framework from prior work~\cite{serval}. To model \earthsight’s distributed execution, we add modules for look-ahead simulation, scheduling, schedule execution, query processing. \revised{Our modifications are limited to the \textit{control logic} governing inference behavior and do not alter underlying mechanics, ensuring the fidelity of our results.}

To evaluate constellation-wide performance, our simulation framework bridges empirical hardware profiling with orbital modeling. Specifically, we execute the filter models on physical Edge TPU and GPU hardware to measure the exact inference latencies and active power draw. We then inject these hardware profiles into the constellation simulator. The simulator tracks the total energy consumed by multiplying the empirically measured hardware power by the simulated workload. Concurrently, the simulator models the satellite's power generation based on solar panel exposure during orbit using SGP4 propagation. By reconciling the physical consumption metrics with the simulated power generation, we can accurately derive systemic metrics such as the percentage of total generated power consumed by the compute payload (Table \ref{tab:power_consumption}) and image delivery latency.

% \revised{This simulation framework incorporates our empirically measured inference times and power profiles to provide accurate scale projections.} For the constellation configuration, we model the PlanetScope Dove network comprising 153 satellites and 14 ground stations, whose publicly available orbital and geographic locations are used for realistic communication and scheduling patterns. The simulation is run over a continuous 48-hour window to evaluate system behavior and stability across varying workloads.
%
\revised{Although our trace-driven simulation methodology cannot model all potential effects of the space environment, such as thermal properties or radiation effects, we note that it is the predominant methodology in all OEC literature~\citep{l2d2, orbitaledge, kodan, serval, 10697471}, due to the difficulty of deploying a constellation of satellites.}

\paragraph{Satellite trajectory and transmission.} Orbits are computed using Two-line-Element descriptors for satellites and propagated using the SGP4 algorithm. Satellites generate power with their solar panels, and power is consumed by the ADACS (control system), camera, receiver, and transmitter. Bandwidth is allocated using the maximum weight independent set algorithm \cite{9003056}, though we found that the choice of routing algorithm negligibly affects latency, provided it is fair. Scheduling and schedule execution take place as described in Sections \ref{system:ground_scheduler} and \ref{system:sat_runtime}.

\paragraph{Baseline.} We use \textsc{Serval} as the baseline~\cite{serval}, referred so in the entire evaluation. We apply the pipeline from Figure~\ref{fig:running_seq}(b) where (1) ground stations communicate the queries, but not a computational schedule like \earthsight, (2) when an image is collected in a region relevant to the queries, the satellite executes the models relevant to those queries. If a query is true, the image is marked as high priority without further inference. If one filter in a term fails, that term's remaining filters are skipped. Unlike our dynamic approach, the baseline (\textsc{Serval}) processes filters statically in order of execution time.

\begin{figure}[t]
    \centering
    \includegraphics[width=0.7\linewidth]{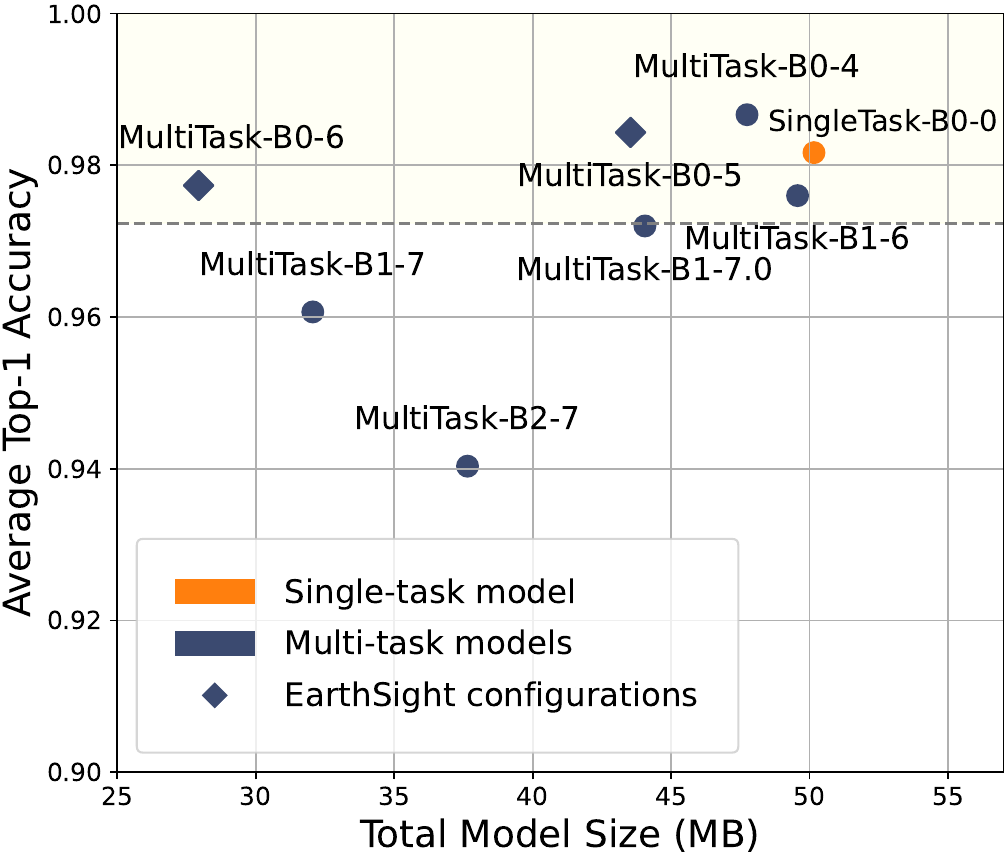}
    \caption{Classification: Model size vs. Top-1 accuracy, across single-task and multi-task configurations. The model names follow the pattern \{model type\}-\{EfficientNet variant\}-\{$N$\}, where layers $1{:}N$ comprise the backbone. }
    \label{fig:configs_classification}
\end{figure}

\subsection{Multi-Task Model Evaluation}
\earthsight's multi-task models dramatically reduce the memory and compute footprint for tasks groups while preserving classification performance, showing that parameter sharing is an effective strategy to reduce computational load on edge devices without degrading accuracy. 

Figure \ref{fig:configs_classification} presents the trade-off between the memory footprint and performance metrics when comparing multiple single-task models with their multi-task counterpart. For classification, the pareto-optimal multi-task configuration cuts memory use by 55.7\% with under 0.5\% accuracy drop, indicating that merging related single-task models into a unified multi-task model preserves model performance. As the backbone grows larger relative to the prediction head, the savings in compute and memory increase, but prediction accuracy suffers as the head has lesser depth to adapt the shared features into the task-specific output. The results for segmentation tasks are similar (Appendix~\ref{app:segmentation_details}); however, their dense nature demands deeper heads than classification.

\subsection{End-to-End Results}

We show \earthsight's globally-aware scheduling with local adaption alleviates the computational bottleneck for prioritizing images, and reduces high-priority image latency across our three different scenarios.

\begin{figure}[H]
    \adjustimage{width=1\columnwidth}{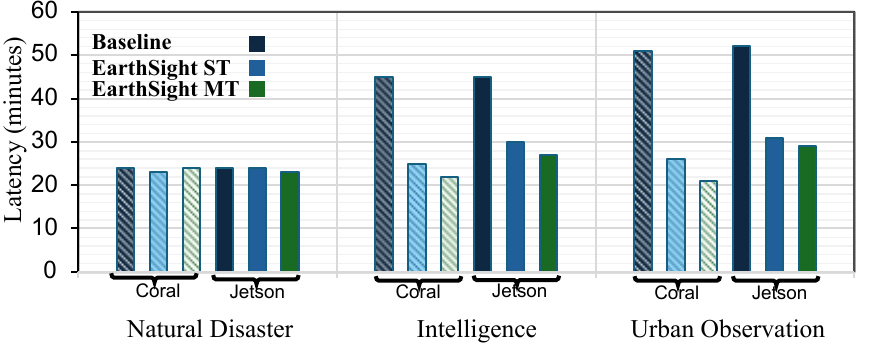}
    \vspace{-2ex}
        \caption{90th percentile tail latency, measured from first ground contact, for high-priority images ($p > 2$) across different scenarios.}

    \label{fig:latency_merged}
\end{figure}
\vspace{-3ex}

\subsubsection{End-to-End Latency}

The ultimate goal of \earthsight's optimized orchestration is to improve the timely delivery of valuable images. Figure \ref{fig:latency_merged} shows the 90th percentile tail latency for high-priority images, measured from the first ground contact made by a satellite after capturing the image, to when the image is downlinked. We compare the baseline, \textsc{Serval}, with \earthsight~using single-task (ST) and multi-task (MT) models with both Coral and Jetson as accelerators.

\paragraph{Performance under High Compute Load (Intelligence Monitoring, Urban Monitoring).} In scenarios characterized by extensive land coverage, a greater number of queries per image, and thus high computational demand, \earthsight~demonstrates significant advantages. As shown in Figure \ref{fig:latency_merged}, all \earthsight~variants substantially reduce tail latency for high-priority images compared to the baseline. For instance, \earthsight~MT using Coral shows a reduction in P90 from 47 to 21 minutes relative to baseline, and 51 to 21 minutes for the Urban Monitoring scenario. This improvement stems from \earthsight's ability to quickly prioritize images and manage the compute queue effectively. The baseline struggles particularly in the Urban scenario when numerous compute-intensive images are captured near ground stations; these images get placed at the rear of the processing queue and miss the downlink window despite their potential value. \earthsight~mitigates this through intelligent processing that nearly eliminates fall-back on first-in first-out image downlink for high-priority images.

\paragraph{Performance under Lower Compute Load (Natural Disaster Scenario).} In the Natural Disaster scenario (See Figure \ref{fig:latency_merged}), the inference workload is lower due to the relative rarity of natural disasters and ease of identification, i.e., smaller models are sufficient for high precision and recall. Thus, despite the large surface area to scan for natural disasters, even the baseline processes most frames successfully, limiting \earthsight's latency improvement. However, as encountering natural disaster images is very rare, the dynamic threshold $\alpha$ is low as only a small percentage of the available bandwidth is used for high-priority images. Because of this, \earthsight~ provides computational and power savings, as shown in Table \ref{tab:power_consumption}. Smarter computation limits energy consumption, which facilitates preparedness in case of sudden increases in load and reduces the battery depth of discharge, a critical component of satellite longevity~\cite{macambira2022}.

\begin{table}[t]
    \centering
    \fontsize{8pt}{9pt}\selectfont
    \caption{Mean Percentage of Generated Power Consumed by Compute during first 6 Hours of Operation.}
    \label{tab:power_consumption}
    \begin{tabular}{lrr}
        \toprule
        \textbf{Approach}& \textbf{TPU} (\%) & \textbf{GPU} (\%) \\ % Units in header
        \midrule
        Baseline & 27.95   & 171.96 \\
        \earthsight~ST   & 11.11   & 69.46   \\
        \earthsight~MT   & \textbf{8.91}    & \textbf{61.49}   \\
        \bottomrule
    \end{tabular}
    \vskip 8pt
\end{table}

\subsubsection{\earthsight's Image Prioritization} 

\begin{table}[t]% htbp: here, top, bottom, page - placement preference
    \centering
    \fontsize{8pt}{9pt}\selectfont
    \caption{Mean and Standard Deviation of Image Prioritization Time (seconds). This benchmarks the filter ordering and pipelining efficiency. Lower is better.}
    \vskip 3pt
    \label{tab:eval_times_grouped}
    \setlength{\tabcolsep}{4pt} % Reduce space between columns
    \begin{tabular}{llrrr}
        \toprule
        Platform & Approach & Mean (s) & Speedup & $\sigma$ (s) \\ 
        \midrule
        \multirow{3}{*}{Jetson} 
            & Baseline         & 2.46 & $1\times$     & 0.95 \\
            & \earthsight~ST    & 1.86 & $1.32\times$ & 0.93 \\
            & \earthsight~MT    & \textbf{1.33} & \textbf{1.85$\times$} & 0.58 \\
        \midrule
        \multirow{3}{*}{Coral} 
            & Baseline         & 7.82 & $1\times$     & 3.00 \\
            & \earthsight~ST    & 5.86 & $1.33\times$ & 2.93 \\
            & \earthsight~MT    & \textbf{4.15} & \textbf{1.88$\times$} & 1.84 \\
        \bottomrule
    \end{tabular}
    \vskip 5pt
\end{table}

\earthsight~accelerates priority assignment to reduce computational load. We measure the time taken from image capture to priority scoring, for images relevant to active queries, measuring the efficiency of \earthsight's filter ordering and pipelining mechanisms. Further analysis of the scheduler’s behavior and theoretical properties is presented in Appendix~\ref{app:theoretical_analysis}.

Table \ref{tab:eval_times_grouped} presents the average time per image required for prioritization by Single-Task (ST) and Multi-Task (MT) versions of \earthsight~compared to the baseline. \earthsight~significantly reduces the mean processing time per image, with the multi-task variant achieving the highest speedup, a reduction in average prioritization time by approximately $1.85\times$ on Jetson and $1.9\times$ on Coral compared to the baseline. This shows optimized evaluation order and avoidance of redundant inference reduce the computational demands of onboard prioritization. The lower standard deviation observed for \earthsight~MT also suggests more consistent processing performance.
\vspace{1ex}

\subparagraph{Comparison with Exact Method.}
Our scheduling policy provides a greedy, approximate solution to minimizing Equation~\ref{eq:objective}. To assess this approximation, we compare it against the exact (exponential-time) optimal solution.

\begin{table}[H]
\centering
\fontsize{8pt}{9pt}\selectfont
\caption{Average evaluation time per image, with Coral coprocessor, on a synthetic suite of 2473 formulas. Lower is better.}
\vskip 5pt
\label{tab:evaluation_times}
\begin{tabular}{lcc}
\toprule
\textbf{Method} & \textbf{Mean Time (s)} & \textbf{Median Time (s)} \\
\midrule
\earthsight~MT & 2.13 & 1.95\\
Exact Solution & 1.98 & 1.95\\
\bottomrule
\end{tabular}
\end{table}

While exact solutions are tractable up to ${\sim}{15}$ filters, scaling to 60 requires roughly a day of compute per formula, despite needing to take milliseconds for effective edge deployment. Thus, we construct a synthetic suite of 2,473 formulas by truncating queries from our three scenarios to contain at most 15 filters. On this benchmark, \earthsight~achieves a relative error of only $7.5\%$ in mean prioritization time compared to the exact method (Table \ref{tab:evaluation_times}). The median times for both approaches are identical, indicating our greedy heuristic aligns with the optimal choice in most cases.

\revised{Although verifying the performance of our filter ordering algorithm relative to the exact solution is not possible beyond 15 filters due to computational intractability, we expect our approximation quality to remain in dense quality workloads. From a theoretical perspective, Appendix \ref{app:theoretical_analysis} demonstrates our utility heuristic is monotone and adaptively submodular, and achieves an expected coverage of at least $(1 - \frac{1}{e})$ the optimal policy within the same onboard budget. It is also typical for approximation quality to improve as the search space grows due to an the increase in near-optimal solutions~\cite{golovin2011adaptive}. Empirically, we saw no performance degradation when scaling from $N=5$ to $N=15$ queries. Irrespective of approximation closeness, \earthsight~still delivers significant improvements over baseline in dense settings (Table \ref{tab:eval_times_grouped}).}

\subsubsection{Ablation Studies}

We conduct ablation studies on the Urban Monitoring scenario to isolate the contribution of key components.

\begin{figure}[H]
    \centering
    \adjustimage{width=1\columnwidth}{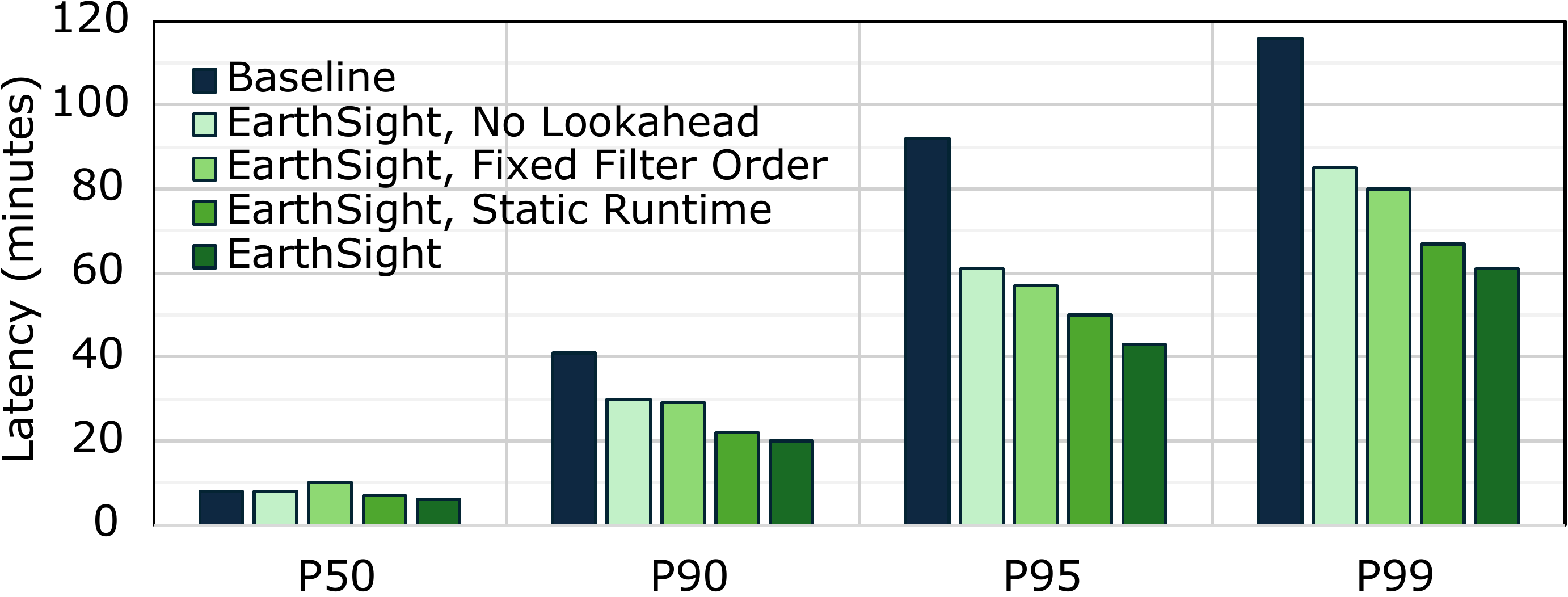}
    \vspace{-2ex}
        \caption{Ablation study results show the impact of individual \earthsight~ components on high-priority ($p>2$) image latency. The combined system performs best.} 
    \label{fig:main_ablation}
\end{figure}

\paragraph{System Components.}
Figure~\ref{fig:main_ablation} compares the full \earthsight~system against baseline and reduced variants where individual components, \emph{optimized filter ordering}, \emph{ground scheduler integration}, and \emph{dynamic thresholding}, are removed. Each component contributes to lowering the latency for high-priority data, with the full \earthsight~configuration achieving the lowest latency across all tasks.

\paragraph{Dynamic Scheduling.} To show the power of the in-orbit runtime's local adaptation, we evaluate the effectiveness of the dynamic priority threshold $\alpha$ compared to static thresholds. Table \ref{tab:strategy_comparison_simple} reveals \earthsight~consistently outperforms all fixed thresholds in mean and P90 latency. The operational context of the satellite is constantly changing. 
Static thresholds are highly sensitive to operational context: if $\alpha$ is set too low, too many low-value images are flagged as high priority, overloading the downlink; if too high, excessive onboard computation delays genuine high-priority data.
By contrast, the dynamic threshold adjusts continuously based on power availability and rejection ratio, maintaining a better balance and yielding lower latency for critical data.

\begin{table}[t]
\centering
\footnotesize
\caption{Comparison of Mean and $P_{90}$ Latency for static and dynamic priority thresholds.}
\label{tab:strategy_comparison_simple}
\begin{tabular}{l rrrrr}
\toprule
Threshold     & Dynamic & 0.2  & 0.3  & 0.4  & 0.5  \\
\midrule
Mean Latency (s)      & \textbf{7.0}  & 11.9 & 7.6  & 8.5  & 10.3 \\
$P_{90}$ Latency (s)   & \textbf{20.0} & 44.7 & 21.1 & 28.4 & 41.2 \\
\bottomrule
\end{tabular}
\end{table}

\begin{table}[t]
\centering
\small
\caption{System metrics when varying model accuracy while holding inference time constant.}
\label{tab:prec_ablation}
\begin{tabular}{cccccc}
\hline
\rule{0pt}{0.8em}
\textbf{Model} & \textbf{Priority} & \textbf{\% High-Pri} & \multicolumn{3}{c}{\textbf{Latency (mins)}} \\
\textbf{Acc.} & \textbf{Acc.} & \textbf{Sent First} & \textbf{P50} & \textbf{P90} & \textbf{P95} \\ \hline

1.00 & 1.00 & 98\% & 7  & 19 & 34 \\
0.95 & 0.97 & 95\% & 7  & 21 & 61 \\
0.90 & 0.93 & 88\% & 8  & 24 & 66 \\
0.80 & 0.83 & 83\% & 8  & 28 & 76 \\
0.60 & 0.70 & 73\% & 10 & 36 & 90 \\ \hline
\end{tabular}
\end{table}

\paragraph{Model Precision and Recall.} We assess how model accuracy affects latency by varying precision/recall while keeping inference time fixed. Table \ref{tab:prec_ablation} shows that as model accuracy falls, both the overall prioritization quality and the percentage of high-priority images sent in the first available downlink slot decline, with modest increases in P90 and P95 tail latencies. Median latency (P50) remains stable as most high-priority images are still correctly prioritized. More severe degradation of tail latency is prevented by the separate priority tier $p_{compute}$ which expedites potentially misclassified images over guaranteed negatives, highlighting system resilience to imperfect model performance.

\paragraph{Robustness under Filter Probability Drift.} \revised{EarthSight is explicitly designed for anomaly discovery, making it resilient to uncertain events. We validated \earthsight’s resilience through perturbing filter pass-probability priors with varying strengths of Gaussian noise ($\sigma$). Results on \earthsight~Multi-Task (Jetson) indicate that even with significant noise, performance degradation is mild. 
As shown in Table~\ref{tab:noise_sensitivity}, latency never exceeds the \earthsight single-task no-noise baseline ($1.9$s) or the standard baseline ($2.6$s), as the utility function continues to select accurate, fast models on average. Estimation errors result only in suboptimal execution order rather than system failure and dropped images.}

\begin{table}[H]
    \centering
    \caption{Prioritization latency statistics (in seconds) under varying Gaussian noise levels ($\sigma$).}
    \label{tab:noise_sensitivity}
    \small
    \begin{tabular}{lccccc}
        \toprule
        \textbf{Noise ($\sigma$)} & \textbf{Mean} & \textbf{Median} & \textbf{Std} & \textbf{Min} & \textbf{Max} \\
        \midrule
        0.00 & 1.33 & 1.04 & 0.52 & 0.64 & 2.98 \\
        0.01 & 1.44 & 1.28 & 0.60 & 0.64 & 3.62 \\
        0.05 & 1.53 & 1.24 & 0.65 & 0.64 & 4.30 \\
        0.10 & 1.55 & 1.30 & 0.64 & 0.64 & 4.17 \\
        0.20 & 1.60 & 1.34 & 0.66 & 0.64 & 4.47 \\
        0.50 & 1.59 & 1.34 & 0.67 & 0.64 & 4.60 \\
        \bottomrule
    \end{tabular}
\end{table}

\section{Related Work}
\label{sec:related-work}
\paragraph{Efficient Machine Learning on the Edge.}
Prior research addresses edge constraints through model compression techniques such as knowledge distillation~\cite{Gou2020KnowledgeDA}, model pruning~\cite{Zhang2022AdvancingMP, fluid}, and weight sharing~\cite{cai2020onceforall}. While these are shown effective in real-time settings such as video analysis and fire detection~\cite{rs15082143, jangirova2025real}, dynamic execution strategies often prove less effective for complex anomaly detection. Early-exit mechanisms~\cite{earlyexit} incur the memory overhead of loading the full backbone, miss deeper, fine-grained features, and have variable latency. Similarly, Mixture-of-Depths models \cite{raposo2024mixtureofdepthsdynamicallyallocatingcompute} introduce storage and runtime decision overheads without overlapping compute. For \earthsight, these overheads outweigh the marginal accuracy gains (0.5–1.0\%) found in Figure \ref{fig:configs_classification}.

\paragraph{Dynamic Scheduling for Satellites.}
\textsc{OEC} was introduced in \cite{orbitaledge} to discard cloudy images identified through onboard compute. \textsc{Kodan}~\cite{kodan} incorporates power availability and geographic location to dynamically trade off execution time for cloud discard and maximize data value density. \textsc{Phoenix}~\cite{liu2024inorbitprocessingnotsunlightaware} schedules inference on sunlit satellites, using inter-satellite links to optimize battery usage across the entire constellation. \textsc{FedSpace}~\cite{so2022fedspace} schedules federated aggregation based on satellite availability to minimize gradient staleness. \revised{\textsc{OrbitChain}~\cite{li2025orbitchainorchestratinginorbitrealtime} distributes compute across multiple satellites on the same orbital track, relying on a theoretical shared-track topology and lacking scale analysis for complex scenarios. These approaches generally address single-task models and do not consider large, multi-purpose constellations.}

\paragraph{Distributing Computation across Ground and Space.}
Hybrid ground-space architectures exploit inexpensive ground computation. \textsc{AdaEO} \cite{10697471} and \textsc{Serval}~\cite{serval} bifurcate queries between the ground stations and satellites, combining prior information with onboard machine learning to determine image utility. This utility informs dynamic image compression and image downlink ordering. We improve on \textsc{Serval} with global context at the ground with local adaptation at the satellite, enabling \earthsight~to scale to multi-purpose, global scenarios.

\paragraph{Stochastic Boolean Function Evaluation.}
The problem our satellite runtime addresses is an NP-hard optimization problem known as Stochastic Boolean Function Evaluation (SBFE)~\cite{deshpande2016approximation}. The objective is to find an evaluation policy that minimizes the expected cost required to determine the truth value of a Boolean function. The outcome of each test (filter) is not known until the cost of running it has been paid. Exact methods based on dynamic programming or value iteration exist but have exponential runtime in the number of tests, rendering them intractable. \revised{Therefore, we use an approximate solution with a submodular cost-benefit heuristic, a common approach that iteratively selects the next test that maximizes objective progress per unit of cost~\cite{golovin2011adaptive}.  While greedy strategies do not guarantee optimality, they have been shown to provide robust empirical performance.}

\balance
\section{Conclusion}

In this work, we show that low-latency, scalable orbital analytics is a \textit{distributing scheduling problem} that requires co-design of globally-aware on-ground planning with adaptive in-orbit execution to conserve limited onboard energy and bandwidth and support diverse real-time scenarios.

Our results reveal \earthsight's distributed decision framework can yield substantial improvements to per-image processing time, power utilization, and end-to-end latency for critical images. \earthsight~easily generalizes to different scenarios and can incorporate auxiliary information from different sources to inform scheduling decisions.

Future work in satellite edge computing may explore mixed-precision frameworks for neural analysis of images or analytics with different stopping points that intelligent runtimes can use to further scale analytical abilities. %Additionally, we recommend optimizing hardware and OS-level decisions to accommodate \earthsight. 

\subsection*{Availability}
We open-source \earthsight. Our results are reproducible with reasonable computational requirements. Please refer to Appendix A.

\subsection*{Acknowledgements}
Special thanks to Irene Wang for her guidance and mentorship during the early stages of this project, Anand Iyer for his oversight as a member of Ansel's undergraduate thesis reading committee, and to our MLSys shepherd Mark Zhao for helping us refine our paper before camera ready. This research was supported in part through cyber-infrastructure resources and services provided by the Partnership for an Advanced Computing Environment (PACE) at the Georgia Techn and gifts from AMD and Google.

% \bibliography{reference}
% \bibliographystyle{mlsys2025}

%%%%%%%%%%%%%%%%%%%%%%%%%%%%%%%%%%%%%%%%%%%%%%%%%%%%%%%%%%%%%%%%%%%%%%%%%%%%%%%
%%%%%%%%%%%%%%%%%%%%%%%%%%%%%%%%%%%%%%%%%%%%%%%%%%%%%%%%%%%%%%%%%%%%%%%%%%%%%%%
% SUPPLEMENTAL CONTENT AS APPENDIX AFTER REFERENCES
%%%%%%%%%%%%%%%%%%%%%%%%%%%%%%%%%%%%%%%%%%%%%%%%%%%%%%%%%%%%%%%%%%%%%%%%%%%%%%%
%%%%%%%%%%%%%%%%%%%%%%%%%%%%%%%%%%%%%%%%%%%%%%%%%%%%%%%%%%%%%%%%%%%%%%%%%%%%%%%
\pagebreak
\appendix

\section*{APPENDIX}
\section{Artifact Appendix}

%%%%%%%%%%%%%%%%%%%%%%%%%%%%%%%%%%%%%%%%%%%%%%%%%%%%%%%%%%%%%%%%%%%%%
\subsection{Abstract}

\earthsight~is a distributed framework that coordinates inference between
ground stations and satellites for low-latency Earth observation. This
artifact includes the full source code of the \earthsight\ satellite
constellation simulator, which extends the Serval simulator~\cite{serval}.
We provide Bash and SLURM scripts that prepare and launch all experiments,
along with Python scripts that reproduce and visualize key results from the
paper. The artifact is publicly available at {\footnotesize\url{https://github.com/scai-tech/EarthSight}} and archived via Zenodo at {\footnotesize\url{https://doi.org/10.5281/zenodo.18826781}}. \earthsight~has been awarded Artifacts Available and Artifacts Functional badges. While some reviewers were able to reproduce our artifact, the reproducibility badge was not awarded due to the prohibitive computational requirements for other reviewers.

%%%%%%%%%%%%%%%%%%%%%%%%%%%%%%%%%%%%%%%%%%%%%%%%%%%%%%%%%%%%%%%%%%%%%
\subsection{Artifact Check-list (Meta-information)}

{\small
\begin{itemize}[nosep]
  \item {\bf Program:} Python-based satellite edge computing simulator.
  \item {\bf Datasets:} All required orbital data and ground-station traces are
        included in the artifact; no external downloads are needed.
  \item {\bf Run-time environment:} Any high-memory Linux CPU node.
        Experiments were conducted on RHEL~9 with Slurm.
  \item {\bf Hardware:} One CPU node with 256\,GB RAM per experiment
        (any modern multi-core CPU is sufficient; memory is the bottleneck).
  \item {\bf Metrics:} End-to-end image delivery latency, per-image compute
        time, and on-board power consumption.
  \item {\bf Output:} Traces, Logs, plain-text result tables and a latency bar chart.
  \item {\bf Disk space required:} $\sim$128\,GB (simulation logs).
  \item {\bf Preparation time:} $\sim$1 hour (environment setup and script
        generation).
  \item {\bf Experiment time:} $\sim$12 hours with a Slurm cluster (jobs run
        in parallel); $\sim$3 days serially for the combined scenario only;
        $\sim$9 days serially for the full suite (18 runs).
  \item {\bf Publicly available:} Yes.
  \item {\bf License:} MIT.
  \item {\bf Archived (DOI):} \url{https://doi.org/10.5281/zenodo.18826781}
\end{itemize}
}

%%%%%%%%%%%%%%%%%%%%%%%%%%%%%%%%%%%%%%%%%%%%%%%%%%%%%%%%%%%%%%%%%%%%%
\subsection{Description}

\subsubsection{How Delivered}
The artifact is available as a public GitHub repository and as a ZIP archive, at the links provided below. All code, reference data, and launch scripts are self-contained.
\begin{itemize}[nosep]
\item Github: \url{https://github.com/scai-tech/EarthSight}
\item Zenodo, with reserved DOI: \url{10.5281/zenodo.18826781.}
\end{itemize}

\subsubsection{Hardware Dependencies}
Experiments were conducted on a node with a singleAMD EPYC 32-core CPU and
256\,GB RAM. \textbf{200+\,GB of RAM is a requirement for each
simulation run}; any modern CPU is otherwise sufficient. If running in parallel, each replica of the program must have its own resources.

\subsubsection{Software Dependencies}
Python 3.9 or 3.13 and scientific Python packages (\texttt{numpy},
\texttt{pandas}, \texttt{skyfield}, \texttt{matplotlib}, \texttt{rtree},
and others) installed from PyPI via \texttt{requirements.txt}.

\subsubsection{Datasets}
All required traces are bundled with the artifact:
\begin{itemize}[nosep]
  \item {\footnotesize\texttt{referenceData/planet\_tles.txt}} --- Planet Labs TLE orbital
        elements.
  \item {\footnotesize\texttt{referenceData/planet\_stations.json}} --- Ground station
        locations.
  \item {\footnotesize\texttt{referenceData/de440s.bsp}} --- JPL planetary ephemeris (32\,MB).
\end{itemize}
Training data for the image classification models is out of scope for this
artifact; filter pass probabilities and execution costs are hard-coded from
the pre-trained models.

%%%%%%%%%%%%%%%%%%%%%%%%%%%%%%%%%%%%%%%%%%%%%%%%%%%%%%%%%%%%%%%%%%%%%
\subsection{Installation}

We recommend Python 3.9 or 3.13. After cloning the repository or extracting
the ZIP file:

\begin{verbatim}

python -m venv ./satsim

# Linux / macOS
source satsim/bin/activate

# Windows (PowerShell)
satsim\Scripts\Activate.ps1

python -m pip install --upgrade pip
python -m pip install -r requirements.txt
\end{verbatim}

Navigate to the simulator directory and verify that all imports succeed:

\begin{verbatim}
cd Sat_Simulator/
python verify_imports.py
\end{verbatim}

%%%%%%%%%%%%%%%%%%%%%%%%%%%%%%%%%%%%%%%%%%%%%%%%%%%%%%%%%%%%%%%%%%%%%
\subsection{Experiment Workflow}

Each simulation runs 48 simulated hours of satellite operation, taking
approximately 12 wall-clock hours per run and requiring 256\,GB RAM.
Tables~4 and~5 are standalone benchmarks that require no simulation data
and can be run immediately after installation. Please refer to the README for commands for easy-to-follow formatting.

\subsubsection{Option A: Slurm Cluster}

All simulation jobs can be submitted in parallel; total wall time is
$\sim$12 hours regardless of suite size.

{\footnotesize
\begin{verbatim}

cd Sat_Simulator

# Generate .sbatch files 
# (replace path, account, and email)
python generate_slurm_scripts.py \
    --cluster-path /path/to/EarthSight \
    --account SLURM_ACCOUNT \
    --email you@institution.edu \
    --combined-only # omit for full suite

cd batch_scripts

# Submit standalone tables immediately 
# (no simulation data needed)
bash launch_tables_standalone.sh

# Submit all simulation jobs in parallel
bash launch_sims.sh

# After all simulation jobs finish:
bash launch_tables_postrun.sh
\end{verbatim}
}

\subsubsection{Option B: Single Machine (No Slurm)}

On Linux or macOS (Windows users: use WSL or Git Bash):

{\footnotesize
\begin{verbatim}
cd Sat_Simulator

# Generate shell scripts
python generate_batch_scripts.py \
    --combined-only
# (omit flag for full 18-job suite)

chmod +x batch_scripts/*.sh \
    batch_scripts/individual/*.sh

# Run standalone tables immediately
bash batch_scripts/generate_table_4.sh 
bash batch_scripts/generate_table_5.sh 

# Run simulations -- choose a scope:
# combined + TPU (~36 h)
bash batch_scripts/run_combined_tpu.sh   
# combined + GPU (~36 h)
bash batch_scripts/run_combined_gpu.sh   
# full combined pipeline (~3 days)
bash batch_scripts/run_all_combined.sh   

# After all simulations finish:
bash batch_scripts/generate_results \ 
        _postrun.sh
\end{verbatim}
}

Individual scripts for each of the 18 simulation configurations are also
provided in \texttt{batch\_scripts/individual/} for fine-grained control.

\subsection{Evaluation and Expected Results}

The above commands, if all are executed, will replicate Tables 3-5 and Figure 6. Please expect up to 5-10\% variability for all experiments. We provide expected outputs in
{\footnotesize\texttt{Sat\_Simulator/results\_expected/}}.

\subsubsection{Tables~4 and~5 --- Compute Time Benchmarks}
These results are produced by standalone scripts and do not depend on
simulation runs. Outputs will be written to
\texttt{results/table4.txt} and \texttt{results/table5.txt}.

Table~4 reports per-image compute time for Serval, \earthsight-STL, and
\earthsight-MTL across Coral/TPU and Jetson/GPU targets, and should reflect
the $\sim$1.9$\times$ speedup reported in the paper. Table~5 compares
\earthsight-MTL against the exact solution and clairvoyant oracle lower bound, showing that \earthsight's multitask scheduling is near-optimal.

\subsubsection{Main Result Figure --- End-to-End Latency}
The bar chart (\texttt{results/summary\_plot.png}) reproduces Figure~3,
showing 90th-percentile end-to-end latency across all scenario and hardware
combinations. Reproduced values should be within $\pm$10\% of the values
reported in the paper; minor variation arises from the stochastic nature of
image capture timing and query-region sampling across simulation runs.

If only the combined (i.e., urban observation) script is ran, then only those results shall be available. The generation script is robust and will simply not generate the bars in the bar chart for simulations that are absent.

\subsubsection{Table~3 --- On-Board Power Consumption}
\textbf{Note on reproducibility.} Table~3 was originally produced from
natural disaster scenario runs of 6 simulated hours, a setting in which
compute demand does not push satellites to their power budget limit. The artifact runs each configuration for the full 48 simulated hours, so the absolute power figures (compute seconds and percentage of generated power consumed) will differ, significantly, from the paper values. Additionally,
\texttt{generate\_table\_3.py} reads only logs whose directory name
contains \texttt{natural} (i.e.\ natural disaster scenario runs); reviewers
who execute only the combined-scenario subset (\texttt{--combined-only})
will observe no output from this script. To reproduce Table~3, at minimum
the natural disaster scenario runs must be completed.

The qualitative claim Table~3 supports---that \earthsight\ operates within
its on-board power budget across hardware platforms---remains verifiable
from the reproduced values regardless of these differences in simulation
length.

\section{Evaluation Scenarios Details}
\label{app:scenarios}

In greater detail, the three scenarios we use for evaluation are below, with supporting sample images from PlanetScope. All images were retrieved from the PlanetScope image gallery: {\footnotesize\url{https://www.planet.com/latest-satellite-imagery-gallery/}}.

\subsection*{Natural Disaster Monitoring}
This scenario provides early warning and impact assessment for events like floods, wildfires, earthquakes, volcanic eruptions, and oil spills. It monitors river water extent, fire thermal anomalies, ground deformation, ash plumes, and the spread of slicks at sea. Prioritization reflects threat urgency, with events like rapidly spreading wildfires or new volcanic eruptions demanding the most immediate attention.

A high-priority task would be triggered by an image showing a rapidly expanding wildfire front, identified by new thermal hotspots adjacent to previously unburned areas, especially when downwind of an urban center. Similarly, an image capturing a new, dense ash plume from a volcano would be flagged for immediate analysis to warn aviation and nearby populations. In contrast, a low-priority task would involve monitoring a known, contained wildfire with a stable perimeter, a steady lava flow far from infrastructure, or observing a river that is swollen but remains within its banks. This tiered approach enables efficient resource allocation that dynamically adapts to threat development.

\begin{figure}[h!]
    \centering
    \includegraphics[width=0.8\columnwidth]{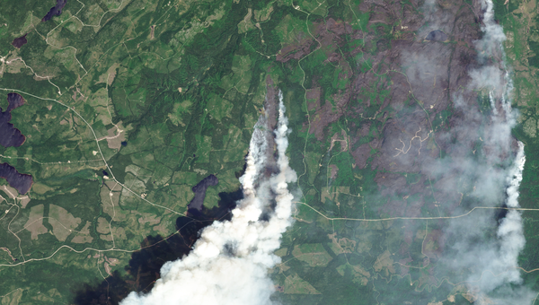}
    \caption{A high-priority image from PlanetScope of a wildfire in California. Onboard analysis can flag new hotspots that signal the fire's imminent spread toward vulnerable areas.}
    \label{fig:wildfire_example}
\end{figure}

\subsection*{Intelligence Scenario}
This scenario delivers strategic and tactical awareness of military and logistical activities in geopolitically significant regions. It monitors key waterways, airbases, and land corridors, detecting anomalous concentrations of hardware and disruptions to critical infrastructure. Continuous operation uses tiered, risk-based priorities to provide timely intelligence on notable patterns for security and diplomatic awareness.

\begin{figure}[h!]
    \centering
    \subfigure[Armament buildup in Yelnya, RU.\label{fig:intel_a}]{
        \includegraphics[width=0.6\linewidth]{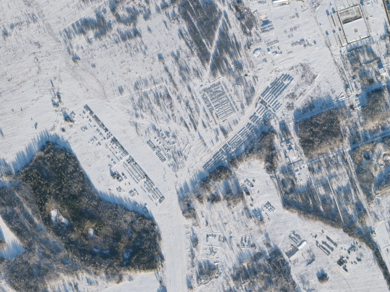}
    }

    \bigskip  % Adds vertical space between the two images

    \subfigure[Destroyed road at Kamaryn-Slavutych.\label{fig:intel_b}]{
        \includegraphics[width=0.6\linewidth]{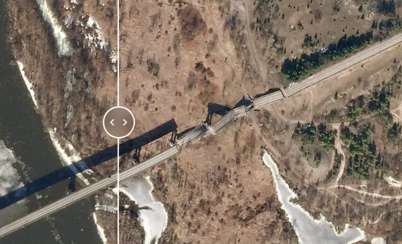}
    }

    \caption{High-priority intelligence images. (a) shows a strategic buildup of military hardware, while (b) shows tactical damage to critical infrastructure.}
    \label{fig:intel_examples}
\end{figure}

For instance, a high-priority event would be the detection of a significant military armament buildup, such as the one observed in Yelnya, Russia, which deviates from baseline activity (Fig.~\ref{fig:intel_a}). Another critical high-priority task is near-real-time damage assessment of key infrastructure, like a destroyed border crossing bridge, indicating conflict escalation or disrupted supply lines (Fig.~\ref{fig:intel_b}). Conversely, an image of a single container ship on a standard shipping route or typical vehicle traffic on an open highway would be a low-priority observation, logged but not prioritized for immediate downlink.

\subsection*{Combined Global Cities Scenario}
This scenario creates a comprehensive monitoring framework for major global cities, integrating disaster, security, and humanitarian elements tailored to city-specific risks. Monitoring adapts to geographic vulnerabilities and population density, addressing complex urban challenges by identifying anomalous patterns in infrastructure and population movement.

A high-priority event in this scenario could be the detection of significant urban flooding in a low-lying coastal city. Another could be the sudden, anomalous massing of vehicles at a key transportation hub, like the Gongabu Bus Terminal in Nepal. Such an event is ambiguous and requires immediate prioritization, as it could signal the start of a mass evacuation (a humanitarian crisis), civil unrest (a security issue), or a major transit failure. Low-priority data would include routine monitoring of city parks for seasonal changes or normal traffic levels at international airports. By fusing these diverse data streams, this scenario provides holistic awareness for densely populated urban centers.

\begin{figure}[h!]
    \centering
    \includegraphics[width=0.8\columnwidth]{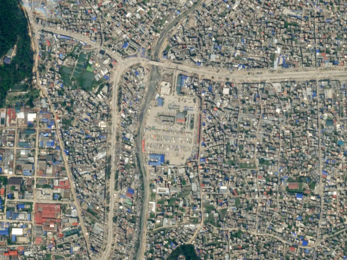}
    \caption{A high-priority image showing an anomalous vehicle buildup at Gongabu Bus Terminal, Nepal. Such an event requires urgent analysis to determine its cause and potential humanitarian or security implications.}
    \label{fig:bus_crowd_example}
\end{figure}

\section{\earthsight's Scheduling Algorithms}
\label{app:pseudocode}

\earthsight's distributed decision platform contains two distinct components. The first component, the ground scheduler, integrates global constellation context in order to prepare a computation schedule for the satellite. The second component, the in-orbit runtime, locally adapts the ground-generated schedule to order filter execution based on resource availability. These components, taken together, achieve a high-performing distributed system that is globally-aware and locally optimal.

\subsection{Ground Station Scheduler}

After the look-ahead simulation is performed, the scheduler has access to the priority threshold $p^*$ for every satellite, which specifies the minimum priority of an image that can be downlinked in the next downlink window.

When a satellite reaches range of a ground station, it requests the computation schedule between the contact and the next communication window. The ground scheduler than prepares the schedule for the satellite according to
Algorithm \ref{alg:ground-schedule}. For each coordinate surrounding which an image will be taken, the algorithm identifies the relevant queries containing the coordinate and synthesizes them into a Disjunctive Normal Form boolean formula $\phi$. The schedule, which is a list of $\phi$ over all coordinates of image capture, is then compressed according to Subsection \ref{system:mitigation} and delivered to the satellite.

\subsection{In-orbit Satellite Scheduler}
On each satellite, the in-orbit satellite runtime dictates how images are processed at a granular, model-execution ordering level. Algorithm \ref{alg:scheduler} provides the pseudo-code for this filter execution engine of the satellite. The execution state $\mathcal{E}$ and confidence $\mathcal{C}_\phi$ are initialized, and the filters are greedily selected for execution based on Equation \ref{eq:utility} until the confidence exceeds the upper threshold $\alpha$ or is less than the lower threshold $\beta$. The values of $\alpha$ and $\beta$ are what enables the local adaptation of the compute schedule, as they are dynamically updated based on the availability of power resources on the satellite.

\begin{algorithm}[tbp]
\caption{Ground Station Schedule Generation}
    \label{alg:ground-schedule}
\begin{algorithmic}[1]
    \Require \( \mathcal{P} \) -- satellite path, \( \mathcal{Q} \) -- query set, \( p^* \) -- priority threshold
    \Ensure \( \mathcal{S} \) -- task schedule, for upcoming satellite pass

    \State Initialize schedule $\mathcal{S} \gets []$

    \For{each planned capture location $\text{loc} \in \mathcal{P}$}
        \State Initialize applicable query set: \( \mathcal{Q}_{\text{loc}} \gets \emptyset \)
    
        \For{each query \( q \in \mathcal{Q} \)}
            \If{\( \text{loc} \in \text{AOI}(q) \)}
                \State \( \mathcal{Q}_{\text{loc}} \gets \mathcal{Q}_{\text{loc}} \cup \{q\} \)
            \EndIf
        \EndFor
        
        \If{\( \mathcal{Q}_{\text{loc}} \neq \emptyset \)}
            \State Compute DNF condition \( \Phi_{\text{loc}} \) over filters from \( \mathcal{Q}_{\text{loc}} \) that yield priority \( \geq p^* \)
        
             \If{{\small \( \mathcal{S} \neq [] \) \textbf{and} last entry in \( \mathcal{S} \) has formula \( \Phi_{\text{loc}} \)}}
                \State Merge \( \text{loc} \) into the last schedule entry
            \Else
                \State Append \( (\text{loc}, \Phi_{\text{loc}}) \) to \( \mathcal{S} \)
            \EndIf
        \EndIf
    \EndFor
    
    \State \Return \( \mathcal{S} \)
\end{algorithmic}
\end{algorithm}

\begin{algorithm}[tbp]
\caption{Satellite Scheduling Runtime}
\label{alg:scheduler}
\begin{algorithmic}[1]
    \Require 
        \(F\) -- Filter set; 
        \(\phi\) -- DNF Boolean formula, over \(F\);
        \(\beta\) -- Lower confidence threshold;
        \(\alpha\) -- Upper confidence threshold
    
    \State Initialize dictionary $\mathcal{E} \gets \{\}$
    \State Initialize $\text{confidence} \gets C_\phi(\mathcal{E})$

    \While{\(\beta \le \text{confidence} \le \alpha\)}
        \State \(f^* \gets \displaystyle \arg\max_{f_i \in F \setminus \mathcal{E}}\,U_{\phi}(f_i, \mathcal{E})\)
        \State Execute \(f^*\) and let \(v^*\) be its Boolean outcome
        \State Update execution state: \(\mathcal{E} \gets \mathcal{E} \cup \{ (f^*, v^*) \}\)
        \State \(\text{confidence} \gets C_\phi(\mathcal{E})\)
    \EndWhile
    
    \State \Return $\text{confidence}$
\end{algorithmic}
\end{algorithm}

\section{Extended Experiments and Vision Task Evaluations}
\label{app:segmentation_details}
Our experiments are conducted on a diverse set of public datasets for both image classification and semantic segmentation. A detailed summary of these datasets, including their resolution, class count, and origin, is provided in Table \ref{tab:dataset_summary_grouped_single}.

Although the \earthsight~pipeline is primarily designed for classification tasks that are utilized to evaluate formulas, we also performed analysis for semantic segmentation tasks. While semantic segmentation models are more complex than necessary for the requirements of prioritization, they can provide richer insights that can aid in immediate response to critical situations.

\begin{figure}[t]
    \centering
    \includegraphics[width=0.7\linewidth]{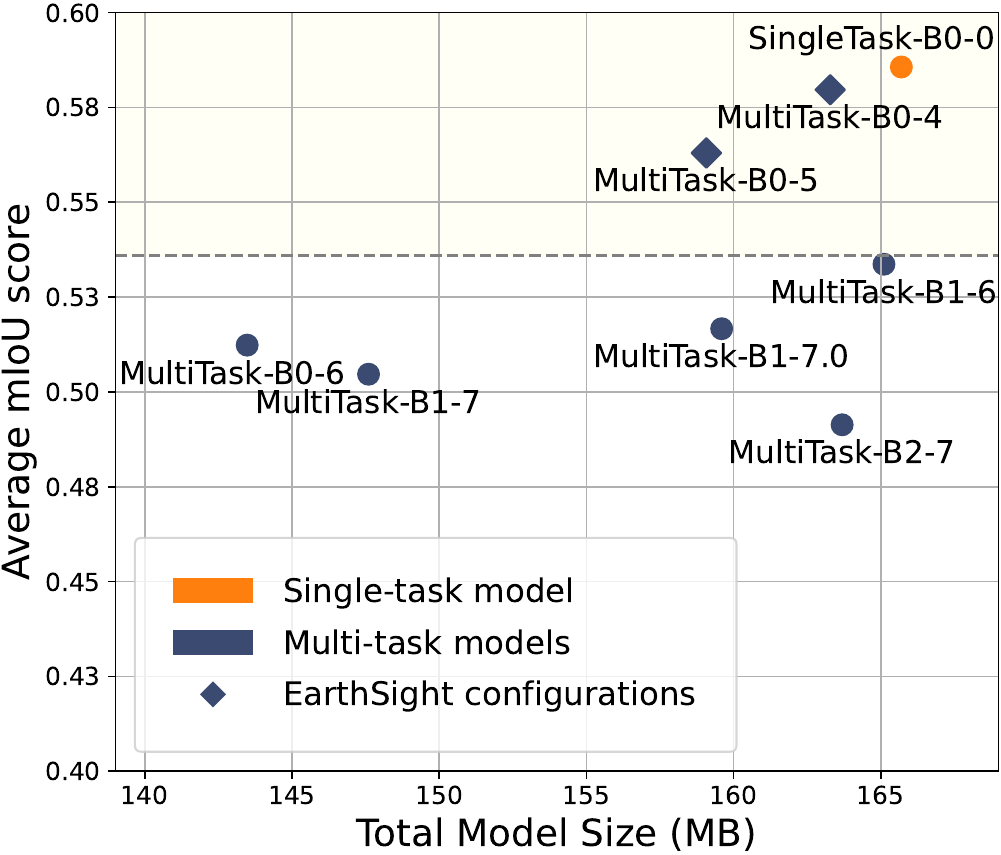}
    \caption{Segmentation: Model size vs. mean Intersection over Union (mIoU).}
    \label{fig:configs_segmentation}
\end{figure}

Our experimental analysis showed that for dense prediction tasks such as semantic segmentation (Figure \ref{fig:configs_segmentation}), it is also viable to reduce the overall model size while maintaining comparable task prediction performance by leveraging a multi-task model paradigm. Since dense prediction requires extracting richer information from input images, employing deeper backbones than EfficientNet-B0, such as ResNet50, or utilizing vision transformer-based feature extractors may lead to more promising results.

\begin{figure*}[t]
\centering
\makebox[0.9\textwidth][c]{%
\includegraphics[width=0.9\textwidth]{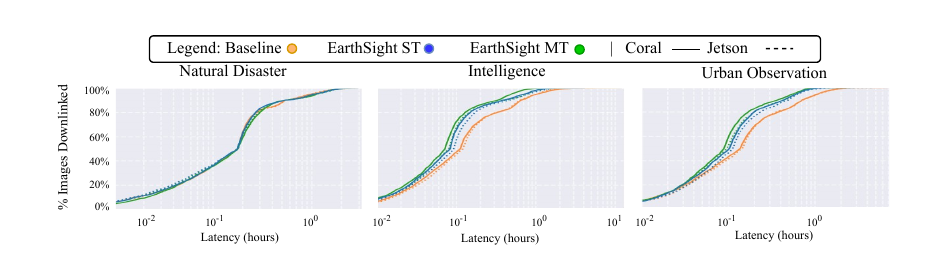}%
}
\caption{Distribution of avoidable image latencies for high priority ($p > 2$) images across all scenarios, measured in hours. \earthsight~shows comparable or improved performance across all scenarios.}
\label{fig:cdfs}
\end{figure*}

Another way to represent the visualizations in Figure \ref{fig:latency_merged} is with a cumulative density function over the images, representing what fraction of the high-priority $(p > 2)$ images are downlinked by time $t$. See \ref{fig:cdfs}. The baseline (orange) appears below both~\earthsight ST and MT, meaning that at any given time, a greater percentage of high-value images would be downlinked by~\earthsight than the baseline approach. This is true for both Coral and Jetson configurations.

\section{Analysis of the Utility Heuristic for Filter Ordering}
\label{app:theoretical_analysis}

\subsection{Comparison to Exact Method and Oracle}

Computing the exact solution minimizing (Eq. \ref{eq:objective}) on even one instance of $\phi$ is intractable for complex formulas, as the runtime is exponential in the number of filters (variables), specifically $O(3^n)$ as each filter is unevaluated, evaluated true, or evaluated false.

To study the performance of the exact solution, we artificially construct formulas from queries that have fewer than 15 filters. For simplicity, we only analyze the performance in the multi-task setting, as our single-task approach is a special case of the multi-task approach.

Lastly, we also compare to an oracle, which has access to the hypothetical results of evaluating every filter on an image, i.e., knows the ground truth. The oracle picks filters minimizing the time (actual, not expected time) to evaluate the formula.

\begin{table}[h!]
\centering
\caption{Average evaluation time per image across different filter ordering methods on a synthetic suite of 2473 formulas. Lower is better.}
\label{tab:evaluation_times_2}
\small
\begin{tabular}{lcc}
\toprule
\textbf{Method} & \textbf{Mean Time (s)} & \textbf{Median Time (s)} \\
\midrule
\earthsight~MT & 2.13 & 1.95\\
Exact Solution & 1.98 & 1.95\\
Oracle         & 1.15 & 1.02\\
\bottomrule
\end{tabular}
\end{table}

On a synthetic suite of 2473 formulas constructed by removing any filters beyond 15 from the existing formulas in our three scenarios, \earthsight~Multi-task achieves an average evaluation time 2.13 seconds, the exact solution achieves an average evaluation time of 1.98 seconds, and the oracle achieves an average evaluation time of 0.85 seconds. See Table \ref{tab:evaluation_times_2} The fact that filter pass probabilities are very low, most often less than 1 to 5\%, contributes to the strong performance of our approximate method. The oracle result demonstrates that the stochastic nature of filter outcomes significantly increases the total compute time from a scheduling perspective.

\subsection{Theoretical Analysis}
The greedy scheduling policy described in Section \ref{system:sat_runtime} aims to find an
approximate solution to the NP-hard Stochastic Boolean Function Evaluation (SBFE) problem
stated in Equation \ref{eq:objective}. While the heuristic is not guaranteed to be optimal,
primarily as it does not look ahead to future costs, its design is theoretically grounded in
the property of adaptive submodularity. This provides strong justification for its quality
and robustness.

The utility function \(U_\phi\) in Equation \ref{eq:utility} maximizes a benefit-cost ratio.
The benefit component, \((1-p_i) \cdot \text{tpr}_i \cdot n_i\), represents the expected
number of DNF terms that will be correctly invalidated by running filter \(f_i\). The quality
of this heuristic can be understood by analyzing the core of this benefit term, \(n_i\).

\newtheorem{theorem}{Theorem}
\begin{theorem}
The benefit function representing the number of unique DNF terms covered by a set of
evaluated filters is a monotone, non-decreasing, and submodular set function.
\end{theorem}

Let \(F\) be the ground set of all filters and \(\phi = \{T_1, \dots, T_m\}\) be the set of
terms in the DNF formula. Define a set function \(g: 2^F \to \mathbb{Z}\) that measures the
benefit of having executed a subset of filters \(S \subseteq F\). Let \(g(S)\) be the number
of terms in \(\phi\) that contain at least one filter from \(S\). Formally:
\[ g(S) = \left| \bigcup_{f \in S} \{ T_j \in \phi \mid f \in T_j \} \right| \]

\paragraph{Monotonicity.} We must show that for any \(A \subseteq B \subseteq F\), we have
\(g(A) \le g(B)\). Let \(\mathcal{T}(S) = \bigcup_{f \in S}\{T_j \in \phi \mid f \in T_j\}\)
denote the set of terms covered by a set of filters \(S\), so that \(g(S) = |\mathcal{T}(S)|\).
If \(A \subseteq B\), then for every \(f \in A\) we also have \(f \in B\), so every term
covered by a filter in \(A\) is also covered by a filter in \(B\); that is,
\(\mathcal{T}(A) \subseteq \mathcal{T}(B)\). Therefore \(g(A) = |\mathcal{T}(A)| \le
|\mathcal{T}(B)| = g(B)\), and the function is monotone non-decreasing.

\paragraph{Submodularity.} We must show that for any subsets \(A \subseteq B \subseteq F\)
and any filter \(f \in F \setminus B\), the following diminishing-returns inequality holds:
\[ g(A \cup \{f\}) - g(A) \ge g(B \cup \{f\}) - g(B) \]

Let \(\mathcal{T}_f = \{T_j \in \phi \mid f \in T_j\}\) be the set of all terms containing
filter \(f\). The marginal gain of adding \(f\) to any set \(S\) with \(f \notin S\) is the
number of terms in \(\mathcal{T}_f\) not already covered by \(S\):
\[ g(S \cup \{f\}) - g(S) = |\mathcal{T}_f \setminus \mathcal{T}(S)| \]

Since \(A \subseteq B\), monotonicity gives \(\mathcal{T}(A) \subseteq \mathcal{T}(B)\).
Therefore:
\[ \mathcal{T}_f \setminus \mathcal{T}(B) \subseteq \mathcal{T}_f \setminus \mathcal{T}(A) \]
and taking cardinalities yields:
\[ |\mathcal{T}_f \setminus \mathcal{T}(A)| \ge |\mathcal{T}_f \setminus \mathcal{T}(B)| \]
which is precisely \(g(A \cup \{f\}) - g(A) \ge g(B \cup \{f\}) - g(B)\), establishing
submodularity.

\paragraph{Implication.}
The submodularity of \(g\) established above justifies the design of our greedy ratio
heuristic, but the correct approximation guarantee depends on how the optimization problem
is stated. The satellite runtime operates under a fixed onboard power budget \(B\), which
is the binding constraint in practice (Section~\ref{system:sat_runtime}). We therefore
state the relevant result in terms of maximization under a budget constraint.

Define \(h(E)\) as the number of DNF terms whose truth value has been \emph{determined}
given execution state \(E\)---either a filter returned \texttt{False}, eliminating the
term, or all filters in the term returned \texttt{True}, satisfying it. The function \(h\)
is monotone and \emph{adaptively submodular} in the sense of Golovin \& Krause
\cite{golovin2011adaptive}: because filter outcomes are independent, conditioning on the
observed state \(E\) does not introduce positive correlations between future outcomes, so
the expected marginal gain of any filter is non-increasing as \(E\) grows. The satellite
runtime's policy---greedily selecting the filter maximizing \(\Delta h(f \mid E) /
t_{\text{eff}}(f, E)\) at each step---is an instance of the adaptive greedy algorithm for
maximizing a monotone adaptive submodular function subject to a cost budget. By Theorem~8
of Golovin \& Krause \cite{golovin2011adaptive}, this policy achieves an expected coverage
of at least \((1 - 1/e) \approx 63\%\) of the coverage attained by the optimal adaptive
policy within the same budget \(B\):
\[ E[h(\pi_{\text{greedy}})] \;\ge\; \left(1 - \frac{1}{e}\right) \cdot
   E[h(\pi^*_B)] \]
where \(\pi^*_B\) is the optimal policy subject to expected cost \(B\). This provides a
formal, constant-factor guarantee for Algorithm~\ref{alg:scheduler} under the
budget-constrained interpretation that reflects the satellite's operational constraints.
We note that for the complementary objective of minimizing cost to reach full
term-resolution, the applicable guarantee is an \(O(\log m)\) approximation (where \(m\)
is the number of DNF terms) via the adaptive stochastic set-cover
result of Deshpande et al.~\cite{deshpande2016approximation}; the empirical gap of 7.5\%
reported in Table~\ref{tab:evaluation_times} suggests the heuristic performs substantially better
than either bound requires in practice.

% \section{Please add supplemental material as appendix here}
% %
% Put anything that you might normally include after the references as an appendix here, {\it not in a separate supplementary file}. Upload your final camera-ready as a single pdf, including all appendices.

%%%%%%%%%%%%%%%%%%%%%%%%%%%%%%%%%%%%%%%%%%%%%%%%%%%%%%%%%%%%%%%%%%%%%%%%%%%%%%%
%%%%%%%%%%%%%%%%%%%%%%%%%%%%%%%%%%%%%%%%%%%%%%%%%%%%%%%%%%%%%%%%%%%%%%%%%%%%%%%

\end{document}